\documentclass{article}

\PassOptionsToPackage{numbers, compress}{natbib}

\usepackage[main, final]{neurips_2026}

\usepackage[utf8]{inputenc} 
\usepackage[T1]{fontenc}    
\usepackage{hyperref}       
\usepackage{url}            
\usepackage{booktabs}       
\usepackage{amsfonts}       
\usepackage{amsmath}        
\usepackage{amssymb}        
\usepackage{nicefrac}       
\usepackage{microtype}      
\usepackage[table]{xcolor}  
\usepackage{graphicx}       
\usepackage{makecell}       
\usepackage{algorithm}      
\usepackage{algpseudocode}  
\usepackage{xspace}         
\usepackage[capitalize]{cleveref} 
\usepackage{multirow}
\usepackage{comment}

\title{DRAMA: Dissecting Attention Redundancy for Accelerating Multimodal Diffusion Large \\ Language Models}

\author{%
  Yi-Hsin Hung$^{1}$, Fangfu Liu$^{2}$, Shao-Yuan Lo$^{1}$, Chu-Song Chen$^{1}$ \\
  $^{1}$National Taiwan University, \, $^{2}$Tsinghua University
 \\
  \texttt{\{r14922a09, sylo, chusong\}@csie.ntu.edu.tw, liuff19@mails.tsinghua.edu.cn} \\
}

\begin{document}

\maketitle

\begin{abstract}
Multimodal diffusion large language models (dLLMs) generate text through iterative masked denoising with bidirectional attention, offering a compelling alternative to autoregressive vision-language modeling. However, each denoising step requires a full forward pass over long multimodal sequences in which visual tokens dominate the sequence length, and a fixed step budget further compounds this quadratic per-step cost.
Existing inference acceleration approaches for multimodal dLLMs characterize attention redundancy implicitly, motivating designs from isolated attention visualizations rather than from measurement.
We close this gap with DRAMA, a training-free acceleration framework grounded in a systematic dissection of attention redundancy in multimodal dLLMs.
By probing attention tensors along spatial and temporal axes, we reveal two persistent redundancy structures: (i) attention mass consistently concentrates on a sparse subset of visual tokens, and (ii) attention distributions rapidly stabilize across denoising steps after an initial formation phase. These findings motivate two complementary mechanisms: Spatial-Redundancy-Aware Visual Token Pruning (SVTP) and Sufficiency-Guided Adaptive Stopping (SGAS). Together, DRAMA achieves an average $18.67\times$ inference speedup across six vision-language benchmarks without training, architectural modification, or attention caching, while maintaining competitive accuracy. Our results establish a strong, measurement-grounded acceleration baseline for multimodal dLLM inference. The code
is available at~\url{https://github.com/CindyHung20/DRAMA}.
\end{abstract}

\section{Introduction}
\label{sec:intro}
Diffusion large language models (dLLMs)~\cite{nie2025large, austin2021structured} generate text by iteratively denoising a fully masked sequence using bidirectional self-attention, enabling parallel token generation and global context reasoning as an alternative to left-to-right autoregressive decoding.
Models such as LLaDA~\cite{nie2025large} and Gemini Diffusion~\cite{geminidiffusion} have demonstrated competitive performance with autoregressive counterparts. Recent extensions, including LLaDA-V~\cite{you2025llada}, LaViDa~\cite{li2025lavida}, and MMaDA~\cite{yang2025mmada}, further bring this paradigm to vision-language tasks by conditioning the denoising process on visual inputs.

While effective, the iterative nature of dLLMs incurs substantial computational cost. This issue is further amplified in multimodal settings.
Each denoising step requires a full forward pass through all $L$ transformer layers over the concatenated sequence $[\mathbf{v};\,\mathbf{c};\,\mathbf{x}_t]$ of visual tokens~$\mathbf{v}$, text condition~$\mathbf{c}$, and noisy output~$\mathbf{x}_t$, where the visual tokens alone contribute 2--3k tokens per image and dominates the total sequence length~$N$.
With a fixed budget of $T$ steps, total cost scales as $\mathcal{O}(T \cdot L \cdot N^2)$, where the visual token count~$N_v$ is the dominant term in~$N$.
Crucially, the visual prefix is never masked and remains static across all steps, yet its quadratic attention cost is repeatedly incurred at every iteration.

\begin{figure}[t]
    \centering
    \includegraphics[width=\linewidth]{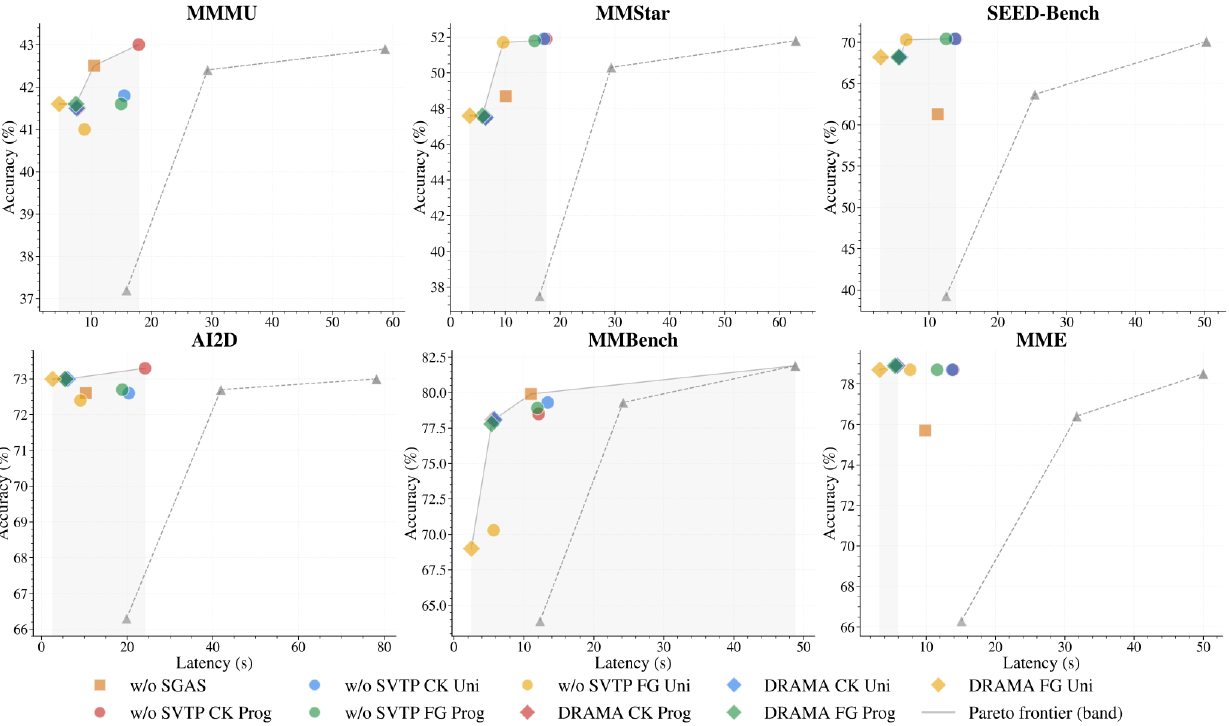}
    \vspace{-0.4cm}
    \caption{\textbf{Accuracy-latency Pareto frontier}. DRAMA variants (diamonds) dominate the Pareto front, achieving higher accuracy at lower latency than all fixed-budget baselines (triangles).}
    \label{fig:pareto}
    \vspace{-0.4cm}
\end{figure}

A concurrent line of work has begun extending acceleration techniques to multimodal dLLMs~\cite{xu2025redvtp, chang2026d3tom, wu2026fast}. Their empirical speedups, together with attention visualizations, suggest the presence of exploitable redundancy in this regime. However, these studies characterize such attention redundancy either implicitly through downstream performance gains or isolated examples. Consequently, their designs are rarely accompanied by a systematic measurement of the attention dynamics they exploit.

In this work, we instead ask how redundancy itself can be systematically exploited to derive an analysis-driven, training-free acceleration framework whose budget is set by measurement rather than by tuning. To answer this, we systematically dissect attention redundancy in multimodal dLLMs along two orthogonal axes: a spatial axis examining how attention mass is distributed over visual tokens, and a temporal axis examining how attention evolves across consecutive denoising steps. Quantified across six vision-language benchmarks, our dissection consistently exposes two analysis-grounded redundancy structures, one per axis.

These findings motivate DRAMA (\underline{D}issecting attention \underline{R}edundancy for \underline{A}ccelerating \underline{M}ultimod\underline{A}l diffusion large language models), a training-free acceleration framework whose two complementary mechanisms each translate one identified redundancy structure into a dedicated strategy. Exploiting the spatial redundancy, Spatial-Redundancy-Aware Visual Token Pruning (SVTP) was designed to retain only dominant visual patches before denoising begins, with no per-step overhead. Sufficiency-Guided Adaptive Stopping (SGAS) exploits the inter-step convergence by terminating denoising once a sufficient decoded sequence has stabilized. By integrating these strategies, DRAMA achieves an average $18.67\times$ inference speedup across six vision-language benchmarks~\cite{kembhavi2016diagram, liu2024mmbench, fu2023mme, yue2024mmmu, chen2024we, li2023seed} on LLaDA-V without any training, architectural modifications, or attention caching, as shown in Tab.~\ref{tab:performance} and Fig.~\ref{fig:pareto}.

Our contributions are summarized as follows:
\begin{itemize}
    \item We close the gap between observing and exploiting attention redundancy in multimodal dLLMs through the first systematic, two-axis dissection of redundancy structures, providing a quantitative basis for deciding what can be removed and how much.
    \item We propose DRAMA, a training-free framework that accelerates multimodal dLLMs via Spatial-Redundancy-Aware Visual Token Pruning and Sufficiency-Guided Adaptive Stopping, grounded in the discovered redundancy structures.
    \item We demonstrate that DRAMA achieves an average $18.67\times$ inference speedup across six benchmarks with competitive performance, providing a strong acceleration baseline for multimodal dLLM inference.
\end{itemize}

\section{Related Work}
\label{sec:related_work}
\noindent\textbf{Diffusion Large Language Models.}
Diffusion large language models (dLLMs) have emerged as an alternative to autoregressive LLMs, offering parallel decoding capabilities and bidirectional attention mechanisms that enable more effective bidirectional reasoning~\cite{nie2025large,song2025seed}. Following the pioneering work of LLaDA~\cite{nie2025large}, subsequent efforts including Mercury Coder~\cite{labs2025mercury}, Gemini Diffusion~\cite{geminidiffusion}, and Seed Diffusion~\cite{song2025seed} have demonstrated the commercial viability of this paradigm. Recent works have extended dLLMs to multimodal settings, including MMaDA~\cite{yang2025mmada}, LLaDA-V~\cite{you2025llada}, and LaViDa~\cite{li2025lavida}.
Despite these advances, dLLMs face several fundamental challenges compared to autoregressive models: a persistent perplexity gap~\cite{arriola2025block, song2025seed}, inability to naturally generate variable-length sequences~\cite{arriola2025block}, and high inference latency due to iterative denoising~\cite{song2025seed}. While dLLMs theoretically offer higher parallelism than autoregressive models, their iterative nature often results in slower practical inference. Our work addresses this gap by proposing interpretable attention analysis-based acceleration strategies specifically designed for multimodal dLLMs.

\noindent\textbf{Inference Acceleration for Diffusion Language Models.}
Existing acceleration methods for dLLMs fall into two categories.
Since bidirectional attention precludes the standard AR KV cache, caching-based approaches~\cite{li2025diffusion, hu2025flashdlm, ma2025dinfer, jiang2025d, ma2025dkv, liu2025dllm, nguyen2025attention, wu2025fast1, song2025sparse, chen2025dpad} reduce redundant computation by reusing intermediate states across denoising steps: dKV-Cache~\cite{ma2025dkv} and dLLM-Cache~\cite{liu2025dllm} adaptively cache key-value or feature states based on token-level dynamics, Elastic-Cache~\cite{nguyen2025attention} triggers layer-aware cache refresh via attention-drift detection, and Fast-dLLM~\cite{wu2025fast} combines block-wise approximate KV caching with confidence-aware parallel decoding.
Sampling-optimization approaches alter the generation procedure itself via parallel~\cite{bao2025learning, huang2025pc, wei2025accelerating, dong2025saber, kongaccelerating, bansal2025enabling} or speculative~\cite{wu2025free, agrawal2025spiffy, gao2025self, hong2025wide} decoding, distillation~\cite{chen2025dparallel, fu2025learnable}, and hybrid modeling methods~\cite{wu2025fast, wang2025diffusion, israel2025planned}.
A concurrent line of work~\cite{xu2025redvtp, chang2026d3tom, wu2026fast} has begun extending acceleration to the multimodal setting. While their empirical gains signal the presence of exploitable redundancy, this redundancy is characterized only implicitly, inferred from downstream speedups or illustrated on isolated visualizations, leaving open how much of it can actually be removed, and where.

\section{Method}
\subsection{Dissecting Spatial-Axis Attention Redundancy}
\label{sec:spatial}
Multimodal dLLMs based on LLaVA-NeXT~\cite{liu2024llavanext} partition each image into $N_p$ patches, yielding 2--3k visual tokens that dominate the sequence length and, consequently, the attention cost at every denoising step. Since this visual prefix is static across all $T$ steps, any spatial redundancy in attention is replicated $T$-fold. We dissect the attention tensors along the spatial dimension at three granularities: individual tokens, image patches, and their cross-step evolution. Each of them reveals a consistent, exploitable structure that generalizes across all evaluated benchmarks. We use MMMU~\cite{yue2024mmmu} for illustration throughout this section, and the full per-benchmark results are in the Appendix~\cref{sec:appendix_redun}.

\begin{figure*}[t]
    \centering
    \includegraphics[width=\textwidth]{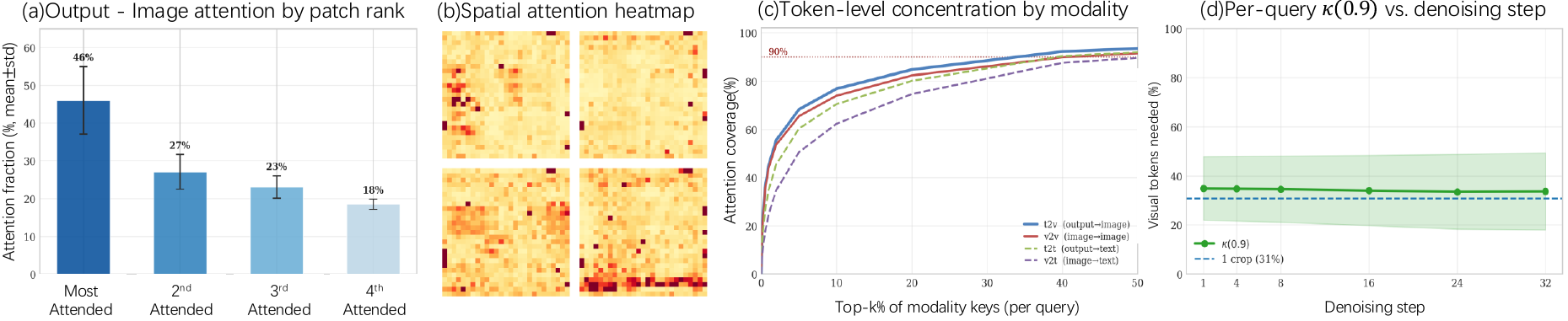}
    \vspace{-0.4cm}
    \caption{\textbf{Spatial redundancy analysis}. \textbf{(a)}~Per-sample attention share of each image patch ranked by dominance. \textbf{(b)}~Per-sample attention share distribution of the most-attended patch across the dataset. \textbf{(c)}~Cumulative attention coverage by top-$k\%$ visual key tokens, stratified by query-key modality pair. \textbf{(d)}~Minimum visual tokens needed for $90\%$ attention coverage ($\kappa(0.9)$) across denoising steps}
    \vspace{-0.2cm}
    \label{fig:spatial}
\end{figure*}

\noindent\textbf{Token-Level Attention Mass Concentration.}
We first examine the finest granularity: the distribution of attention mass over individual visual key tokens, stratified by the modality of query-key interaction. For each query token, we sort its attention weights over the modality-specific key set and compute cumulative coverage at varying retention fractions (\cref{fig:spatial}(c)). The two cross-modal pairs involving visual keys, namely output$\to$image (\textbf{t2v}) and image$\to$image (\textbf{v2v}), exhibit strikingly concentrated profiles: the top $1\%$ of visual key tokens already capture ${\sim}47\%$ of a query's attention mass, and the top $10\%$ suffice for ${\sim}78\%$ coverage.
By contrast, text-key interactions (\textbf{t2t} and \textbf{v2t}) follow a markedly flatter curve, with the top $1\%$ accounting for only ${\sim}26\%$ and ${\sim}33\%$, respectively.
This asymmetry confirms that the extreme sparsity is specific to the visual modality: the model concentrates its attention budget on a small set of visually salient key positions while treating the remaining visual tokens as near-irrelevant context.

\noindent\textbf{Patch-Level Non-Uniformity.}
If salient tokens were spread uniformly across patches, each patch would receive ${\sim}1/N_p$ of the output$\to$image attention, leaving no efficient patch-level pruning strategy.
Instead, the distribution is highly skewed within individual samples: the most-attended patch captures ${\sim}44\%$ of aggregate attention, which is $1.7\times$ the uniform baseline, and the top two patches suffice for ${\sim}68\%$ coverage, while the remaining patches contribute marginally (\cref{fig:spatial}(a)).
Patches are ranked per-sample before averaging, so these fractions reflect genuine per-sample concentration rather than a cross-sample aggregation artefact. The dominant patch's attention share is consistent across the dataset (see Appendix~\cref{sec:appendix_spatial}), confirming a stable structural property of the model.
This cross-patch skewness implies the effective visual receptive field is far smaller than the full token set, motivating our spatial-redundancy-aware visual token pruning (\cref{sec:VTP}).

\noindent\textbf{Step-Invariant Spatial Redundancy.}
A critical question for acceleration design is whether the spatial redundancy structure changes as denoising progresses. We track the per-query metric $\kappa_i(0.9)$, the minimum number of visual key tokens needed to capture $90\%$ of an output query's attention mass, averaged across all layers and queries at each denoising step (\cref{fig:spatial}(d)). 
Across all benchmarks, $\kappa(0.9)$ remains remarkably stable: it accounts for only ${\sim}27\%$ of the total visual token budget at every step, with a relative variation below $5\%$ from the first step to the last. The near-flat trajectory demonstrates that the degree of spatial concentration is step-invariant. Moreover, the identity of the salient positions is also preserved: the inter-step attention divergence measured in \cref{sec:temporal} drops to near zero during the plateau phase, and Pinsker's inequality bounds each token's weight change accordingly (formal derivation in~\cref{sec:stopping}), confirming that a patch selection made before denoising remains valid across all $T$ steps.

\subsection{Dissecting Temporal-Axis Attention Redundancy}
\label{sec:temporal}
The spatial analysis dissects redundancy across tokens, identifying which visual positions can be pruned. Here, we dissect redundancy across denoising steps, examining whether the full attention distribution changes enough between consecutive steps to justify repeating the $\mathcal{O}(L \cdot N^2)$ forward pass. When it does not, successive steps recompute near-identical outputs at full cost, constituting temporal redundancy.
We quantify this via the divergence of per-query attention distributions at consecutive steps across the full trajectory.
\noindent\textbf{Inter-Step Distributional Divergence.}
Following~\cite{pu2024efficient}, we adopt the Jensen-Shannon divergence (JSD),  a symmetric, $[0, \ln 2]$-bounded variant of KL divergence, as our inter-step attention redundancy metric to measure how attention distributions change between consecutive steps.
For each layer $l$, head $h$, and query position $i$, the attention weights at denoising step $t$ form a probability distribution over all $N$ key positions:
\begin{equation}
    \mathbf{a}^{(t)}_{l,h,i} = \mathrm{Softmax}\!\left(\frac{\mathbf{q}^{(t)}_{l,h,i} \cdot \mathbf{K}^{(t)\top}_{l,h}}{\sqrt{d_k}}\right) \in \Delta^{N-1},
    \label{eq:attn_dist}
\end{equation}
where $\Delta^{N-1}$ denotes the probability simplex.
The JSD between consecutive-step attention distributions at the same query position is
\begin{equation}
    \mathrm{JSD}\!\left(\mathbf{a}^{(t)} \,\big\|\, \mathbf{a}^{(t-1)}\right) = \tfrac{1}{2}\,\mathrm{KL}\!\left(\mathbf{a}^{(t)} \,\big\|\, \mathbf{m}\right) + \tfrac{1}{2}\,\mathrm{KL}\!\left(\mathbf{a}^{(t-1)} \,\big\|\, \mathbf{m}\right),
    \label{eq:jsd}
\end{equation}
where $\mathbf{m} = \tfrac{1}{2}(\mathbf{a}^{(t)} + \mathbf{a}^{(t-1)})$ is the midpoint distribution.
Unlike \cite{pu2024efficient}, which compares attention across spatial positions within a single step, we compare the same (layer, head, query) triplet across two consecutive steps.
We aggregate into a mean per-layer divergence and its normalized redundancy complement:
\begin{equation}
    \bar{J}_l(t) = \frac{1}{H \cdot N} \sum_{h=1}^{H} \sum_{i=1}^{N} \mathrm{JSD}\!\left(\mathbf{a}^{(t)}_{l,h,i} \,\big\|\, \mathbf{a}^{(t-1)}_{l,h,i}\right),
    \qquad
    R_l(t) = 1 - \frac{\bar{J}_l(t)}{\ln 2}.
    \label{eq:mean_jsd}
\end{equation}
$R_l(t) \!\to\! 1$ indicates near-identical consecutive distributions.
\begin{figure*}[t]
    \centering
    \includegraphics[width=\textwidth]{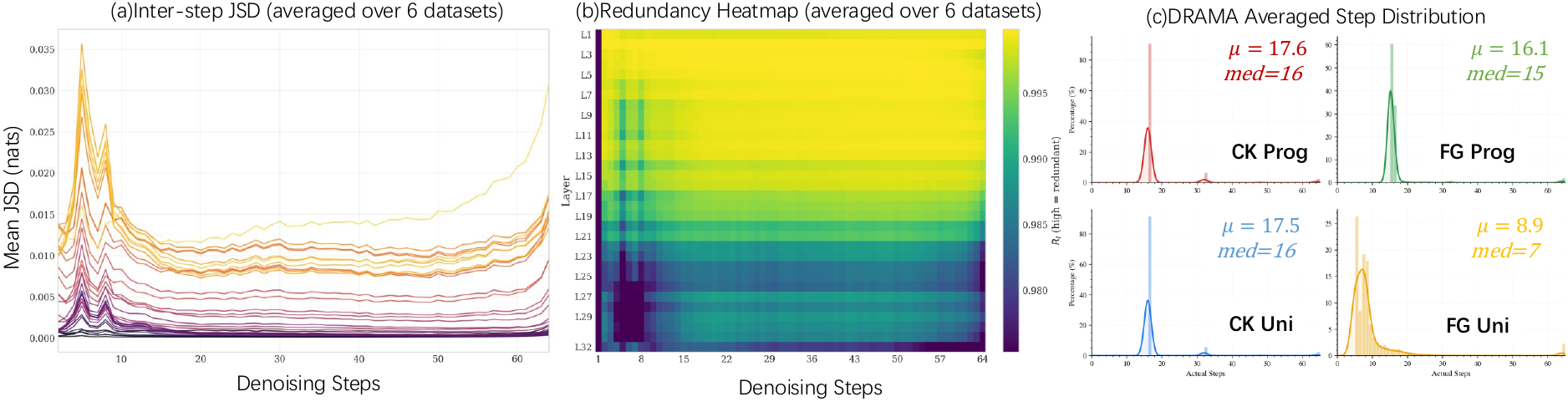}
    \vspace{-0.4cm}
    \caption{\textbf{Temporal redundancy analysis across the denoising trajectory} (averaged over six benchmarks). \textbf{(a)}~Per-layer inter-step JSD ($\bar{J}_l$) between consecutive denoising steps; peaks highlight steps of greatest attention change. \textbf{(b)}~Redundancy heatmap $R_l(t)$ across layers and steps; bright regions indicate high redundancy ($R_l \!\to\! 1$). \textbf{(c)}~Distribution of actual stopping steps of our proposed DRAMA, whose concentration aligns with the plateau onset in~(a).}
    \vspace{-0.1cm}
    \label{fig:temporal}
\end{figure*}

\noindent\textbf{Phase-Structured Denoising Dynamics.}
\cref{fig:temporal}(a) plots $\bar{J}_l(t)$ across the full denoising trajectory for each transformer layer, averaged over all evaluation benchmarks.
Rather than declining monotonically, the inter-step divergence reveals a characteristic three-phase temporal structure:

\noindent\emph{(i) Structure formation} ($\bar{J}_l$ peak)\emph{.}
The JSD peaks sharply as the sequence transitions from fully masked to carrying initial content.
Although only ${\sim}N_g/T$ tokens are revealed per step, they represent the first semantic signal in an otherwise content-agnostic attention landscape. Naturally resulting in a maximized relative information gain and inducing the largest attention redistribution in the entire trajectory.

\noindent\emph{(ii) Stable plateau} ($\bar{J}_l \!\to\! 0$)\emph{.}
Once sufficient context has been established, each subsequent unmasking becomes marginal, and the redundancy score rises to $R_l(t) > 0.99$ across all layers (\cref{fig:temporal}(b), yellow region).
Because $\mathbf{O}^{(t)}_l = \mathbf{A}^{(t)}_l \mathbf{V}^{(t)}_l$, near-identical attention maps ($\mathbf{A}^{(t)}_l \!\approx\! \mathbf{A}^{(t-1)}_l$) combined with only marginally updated values indicate approximately stationary outputs $\mathbf{O}^{(t)}_l \!\approx\! \mathbf{O}^{(t-1)}_l$ during this denoising phase.
This plateau spans the majority of the trajectory and constitutes the primary locus of temporal redundancy: the model pays full $\mathcal{O}(L \cdot N^2)$ cost per step for diminishing informational return.

\noindent\emph{(iii) Terminal refinement} ($\bar{J}_l$ secondary rise)\emph{.}
A secondary JSD rise emerges as each remaining step must resolve a growing fraction of the remaining ambiguity: with ${\sim}N_g/T$ tokens unmasked per step but progressively fewer masked positions left, each unmasking carries disproportionate semantic weight, requiring fine-grained disambiguation that redistributes attention in the deeper layers.

\subsection{DRAMA}
\begin{figure*}[t]
    \centering
    \includegraphics[width=\textwidth]{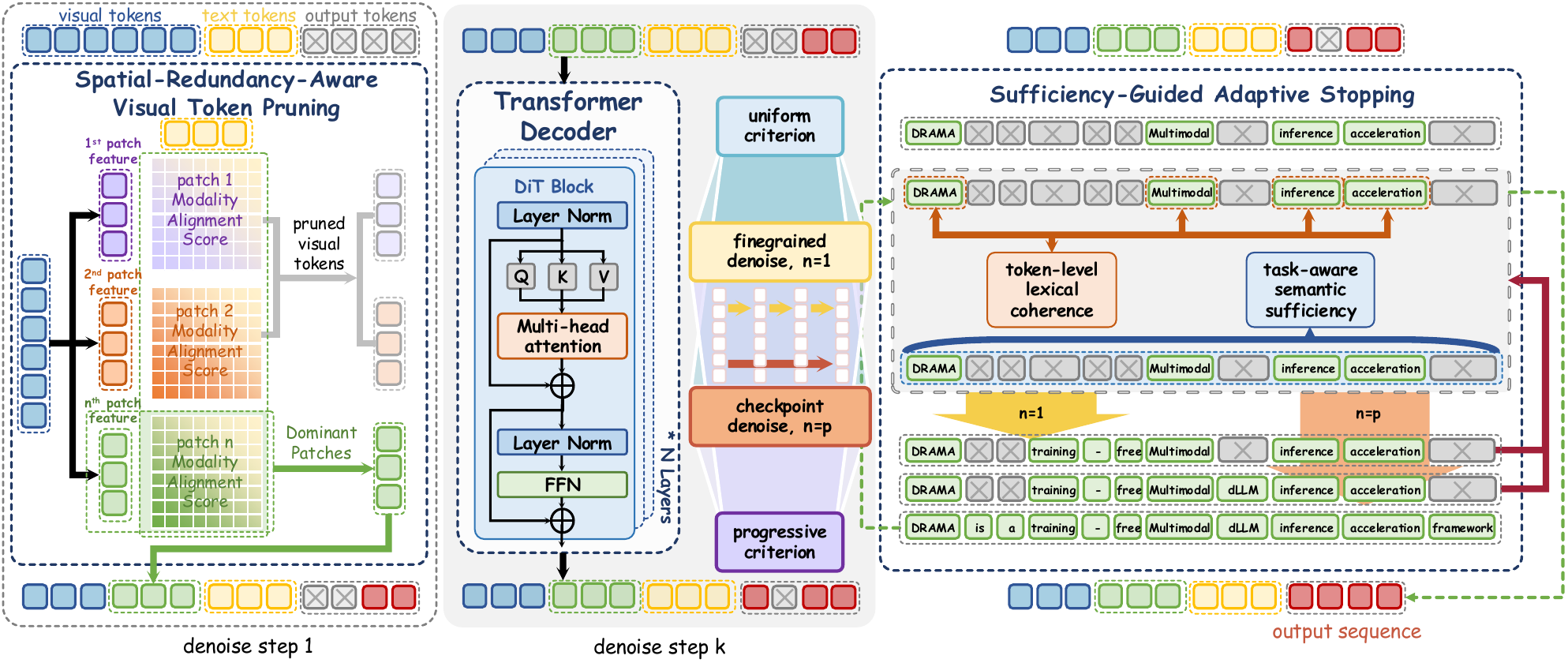}
    \vspace{-0.4cm}
    \caption{\textbf{Overview of DRAMA.} DRAMA applies Spatial-Redundancy-Aware Visual Token Pruning (SVTP) to select a compact patch subset once before denoising. Sufficiency-Guided Adaptive Stopping (SGAS) terminates the denoising trajectory early.}
    \vspace{-0.1cm}
    \label{fig:pipeline}
\end{figure*}
\subsubsection{Spatial-Redundancy-Aware Visual Token Pruning.} 
\label{sec:VTP}
Given the extreme spatial concentration identified in \cref{sec:spatial}, a natural strategy is to retain only the most attended visual tokens at each denoising step. However, scoring by attention from the current step requires a forward pass that has already been computed. A cross-step reuse variant that scores tokens using the previous step's attention would circumvent this, but still introduces per-step overhead across all $T \!\cdot\! L$ layer passes and faces a cold-start problem at $t\!=\!1$.
Crucially, both strategies overlook a key finding from our analysis: the identity of the dominant patches is content-dependent yet step-invariant (\cref{sec:spatial}), meaning a single, well-chosen selection made before denoising can remain valid across all $T$ steps. This motivates a fundamentally different approach: rather than relying on attention signals that are either unavailable or redundantly recomputed, we score patches by their cross-modal relevance to the input question, performing the selection at the beginning of the denoising process at zero marginal cost.

Specifically, the frozen vision encoder and the dLLM's token embedding layer project visual patches and question tokens into a shared representation space, providing a natural relevance signal without any additional computation. We score each patch $i$ by the similarity between its mean-pooled embedding $\bar{\mathbf{p}}_i$ and the mean-pooled question representation $\bar{\mathbf{q}}$ in this space, and construct the active set $\mathcal{P}$ by retaining the global low-resolution overview base patch together with the highest-scoring sub-patches. At every denoising step $t$ and layer $l$, attention is computed over the filtered key-value set:
\begin{equation}
    \hat{\mathbf{A}}^{(t)}_l
    = \mathrm{Softmax}\!\left(
      \frac{\mathbf{Q}\, \mathbf{K}[\mathcal{P} \cup \mathcal{T}]^\top}{\sqrt{d_k}}
    \right)
    \mathbf{V}[\mathcal{P} \cup \mathcal{T}],
    \;\;\text{where}\;\;
    \mathcal{P} \!=\! \{0\} \cup \bigl\{i \mid s_i \!\geq\! s_{(|\mathcal{P}|)}\bigr\},
    \label{eq:pruned_attn}
\end{equation}
with $s_i = \mathrm{sim}(\bar{\mathbf{p}}_i, \bar{\mathbf{q}})$ and $\mathcal{T} = \mathcal{I}_c \cup \mathcal{I}_g$ denoting all text token indices.

This reduces the effective key-value length from $N$ to $|\mathcal{P}| \!\cdot\! M + |\mathcal{T}|$; since visual tokens constitute over $90\%$ of the sequence ($N_v \gg N_c + N_g$), even moderate pruning yields a substantial per-step speedup that compounds across all $T$ steps.
The entire scoring procedure involves only $N_p$ inner products in $\mathbb{R}^D$ and is performed once before the first denoising step, introducing no per-step overhead or attention caching.
\subsubsection{Sufficiency-Guided Adaptive Stopping.}
\label{sec:stopping} 
The plateau established in \cref{sec:temporal} shows that, beyond the structure-formation phase, consecutive denoising steps produce near-identical attention distributions across all layers. A Pinsker-type bound on the resulting layer-output drift (full derivation in App.~\ref{app:bound}) then propagates this stationarity through the residual stream and the output projection to the decoded token sequence, allowing us to elevate the decoded text to a token-space sufficiency surrogate. This surrogate supplants any runtime monitoring of $\bar{J}_l(t)$, which would otherwise re-introduce the per-layer attention caching we seek to eliminate, with a stopping rule defined in decoded sequence space.

\noindent\textbf{Stop predicate.}
We frame the stopping rule as the first-hit time of a conjunction of two failure-mode-orthogonal predicates,
\begin{equation}
\begin{gathered}
    t^{\star} =\; \min\bigl\{\, t \geq T_{\min} : t \in \mathcal{E},\; \mathrm{Suff}(\mathbf{x}_t) \wedge \mathrm{Coh}(\mathbf{x}_t) \,\bigr\}, \\
    \text{where}\;\; \mathrm{Suff}(\mathbf{x}_t) :=\; \mathbb{1}\bigl[\textsc{Parse}(\mathbf{x}_t) \in \mathcal{V}_{\textit{task}}\bigr],  \,\,
    \mathrm{Coh}(\mathbf{x}_t):=\; \mathbb{1}\bigl[\boldsymbol{r}(\mathbf{x}_t) \in \mathcal{R}_{\boldsymbol{\theta}(t)}\bigr],
\end{gathered}
\label{eq:stop_pred}
\end{equation}
with $\boldsymbol{r}(\mathbf{x}_t) = (r^{\mathrm{div}}, r^{\mathrm{rep}}, r^{\mathrm{con}})$ the diversity, peak-frequency, and consecutive-repetition statistics of $\mathbf{x}_t$, $\mathcal{R}_{\boldsymbol{\theta}(t)}$ their admissible region under the threshold schedule $\boldsymbol{\theta}(t)$, $\mathcal{V}_{\textit{task}}$ the set of structurally valid task answers, and $\mathcal{E}$ the granularity-controlled evaluation set; the procedure returns $\mathbf{x}_{t^{\star}}$, defaulting to the full-budget output $\mathbf{x}_T$ when the predicate never triggers, so the worst-case behaviour gracefully recovers the unaccelerated baseline. The two predicates discharge complementary failure modes: $\mathrm{Suff}$ alone admits parseable answers embedded in degenerate near-mask sequences, while $\mathrm{Coh}$ alone admits fluent but task-irrelevant continuations, so their conjunction rules out both pathologies by construction, requiring neither retraining nor an auxiliary verifier. The concrete forms of $\textsc{Parse}$, $\boldsymbol{r}$, and the per-variant schedules $\boldsymbol{\theta}(t)$ are detailed in App.~\ref{sec:appendix_sgas} due to space.

\noindent\textbf{Design space.}
Eq.~\eqref{eq:stop_pred} exposes two orthogonal design axes whose Cartesian product instantiates the 2$\times$2 design space ablated in \cref{sec:ablation}. Evaluation granularity controls when the predicate is queried along the trajectory: a \textit{checkpoint} schedule evaluates only on a sparse set $\mathcal{C}$, amortising decoding overhead over a coarse partition, whereas a \textit{fine-grained} schedule evaluates at every step beyond $T_{\min}$ to admit the earliest viable stop. Threshold scheduling controls how strictly the predicates are evaluated: a \textit{progressive} schedule tightens $\boldsymbol{\theta}(t)$ early and relaxes it later, encoding the structure-formation prior surfaced in \cref{sec:temporal}, whereas a \textit{uniform} schedule applies fixed thresholds throughout, encoding the absence of such a prior. The two axes are independent by construction, with the first governing temporal density and the second semantic strictness, so the four resulting variants form a 2$\times$2 instantiation whose two axes correspond to the two redundancy structures measured in Section~\ref{sec:temporal} and~\ref{sec:spatial}. The full procedure (\cref{alg:stopping}, deferred to App.~\ref{sec:appendix_sgas}) and the empirical stopping distribution under each variant (\cref{fig:step_dist}) concentrate near the plateau onset predicted in \cref{sec:temporal}, closing the loop between dissection and design.

\section{Experiments}
\label{sec:experiments}

\subsection{Experimental Setup}
\definecolor{rank1}{HTML}{E0C0FC}
\definecolor{rank2}{HTML}{F0E0FE}
\definecolor{abrank1}{HTML}{D0D0D0}
\definecolor{abrank2}{HTML}{E8E8E8}
\definecolor{hdrBlue}{HTML}{D4E5FC}
\definecolor{hdrRed}{HTML}{F5D0CF}
\definecolor{hdrGold}{HTML}{FCF0CC}
\definecolor{hdrGreen}{HTML}{CEEDD8}
\definecolor{hdrOrange}{HTML}{F5DBC5}
\noindent\textbf{Datasets and Baselines.}
We evaluate on six vision-language benchmarks spanning diverse task formats:
AI2D~\cite{kembhavi2016diagram}, MMBench~\cite{liu2024mmbench}, MME~\cite{fu2023mme}, MMMU~\cite{yue2024mmmu}, MM-Star~\cite{chen2024we}, and SEED-Bench~\cite{li2023seed}. For SEED-Bench, we evaluate on a randomly sampled subset of 2,000 samples on LLaDA-V and 1,000 samples on LaViDa, following LLaDA-V's own protocol; all other benchmarks are evaluated in full. We compare against LLaDA-V~\cite{you2025llada} with fixed denoising budgets of $T \!\in\! \{16, 32, 64\}$ steps, representing the natural baseline of uniform step reduction. The 64-step configuration serves as the unaccelerated reference throughout. We further provide comparisons of LaViDa on five of the evaluation benchmarks, in Appendix~\cref{tab:performance_lavida}.

\noindent\textbf{Implementation Details.}
We adopt LLaDA-V~\cite{you2025llada}, a 7B-parameter multimodal dLLM built on $L\!=\!32$ transformer layers with $H\!=\!32$ attention heads and a frozen SigLIP2~\cite{tschannen2025siglip} vision encoder, as our evaluation backbone.
Inference is performed in bfloat16 on a single NVIDIA RTX PRO 6000 GPU (96\, GB) with a batch size of~1.
Generation uses a single block of $N_g\!=\!128$ tokens with deterministic decoding (temperature~$0.0$) and low-confidence remasking~\cite{nie2025large}. For SGAS, the maximum denoising budget is $T\!=\!64$ with $T_{\min}\!=\!1$; checkpoint mode evaluates at steps $\mathcal{C}\!=\!\{16, 32, 48, 64\}$, and fine-grained mode evaluates at every step beyond $T_{\min}$. The full study consumes $747.99$\, GPU-hours in total (including baselines, ablations, and preliminary runs), of which $82.18$\, GPU-hours are spent on the four reported DRAMA variants.
\begin{table*}[!t]
\renewcommand{\arraystretch}{0.9}
\centering
\caption{Results on six benchmarks. Metrics: Acc (\%, $\uparrow$), Latency (s/sample, $\downarrow$), TPS (tokens/s, $\uparrow$), Speedup vs.\ Baseline 64 steps ($\uparrow$). w/o SGAS uses 32 denoising steps; its Speedup is normalized by $\div 2$ to discount the inherent gain from halving the step count. \colorbox{rank1}{\strut Best}\ \colorbox{rank2}{\strut 2nd best}\ (Baselines and DRAMA); \colorbox{abrank1}{\strut Best}\ \colorbox{abrank2}{\strut 2nd best}\ (Ablation). $^{\dagger}$Evaluated on a subset of the full benchmark.
}
\label{tab:performance}
\resizebox{\textwidth}{!}{%
\begin{tabular}{l l c c c c c c c c c c c c}
\toprule
\multicolumn{2}{c}{} & \multicolumn{3}{c}{\textbf{Baseline}} & \multicolumn{5}{c}{\textcolor{gray}{\textbf{Ablation}}} & \multicolumn{4}{c}{\textbf{DRAMA (Ours)}} \\
\cmidrule(lr){3-5} \cmidrule(lr){6-10} \cmidrule(lr){11-14}
 & \textbf{Metric}
& \makecell[c]{\footnotesize 64 \\ \footnotesize steps}
& \makecell[c]{\footnotesize 32 \\ \footnotesize steps}
& \makecell[c]{\footnotesize 16 \\ \footnotesize steps}
& \makecell[c]{\textcolor{gray}{\footnotesize w/o} \\ \textcolor{gray}{\footnotesize SGAS}}
& \makecell[c]{\textcolor{gray}{\footnotesize w/o SVTP} \\ \textcolor{lightgray}{\footnotesize FG Prog}}
& \makecell[c]{\textcolor{gray}{\footnotesize w/o SVTP} \\ \textcolor{lightgray}{\footnotesize FG Uni}}
& \makecell[c]{\textcolor{gray}{\footnotesize w/o SVTP} \\ \textcolor{lightgray}{\footnotesize CK Prog}}
& \makecell[c]{\textcolor{gray}{\footnotesize w/o SVTP} \\ \textcolor{lightgray}{\footnotesize CK Uni}}
& \makecell[c]{\footnotesize FG Prog}
& \makecell[c]{\footnotesize FG Uni}
& \makecell[c]{\footnotesize CK Prog}
& \makecell[c]{\footnotesize CK Uni} \\
\midrule
\addlinespace[3pt]
\rowcolor{gray!12}\multicolumn{14}{l}{\textbf{AI2D}} \\
\addlinespace[2pt]
 & Acc (\%) & \cellcolor{rank1}\textbf{73.0} & 72.7 & 66.3 & 72.6 & 72.7 & 72.4 & \cellcolor{abrank1}\textbf{73.3} & 72.6 & 73.0 & \cellcolor{rank1}\textbf{73.0} & \cellcolor{rank2}73.0 & \cellcolor{rank2}73.0 \\
\addlinespace[2pt]
 & Latency (s) & 78.2 & 41.9 & 19.9 & \cellcolor{abrank2}10.4 & 18.9 & \cellcolor{abrank1}\textbf{9.1} & 24.2 & 20.4 & \cellcolor{rank2}5.5 & \cellcolor{rank1}\textbf{2.6} & 5.7 & 6.1 \\
\addlinespace[2pt]
 & TPS (tok/s)& 2.4 & 4.2 & 9.2 & \cellcolor{abrank2}12.2 & 9.8 & \cellcolor{abrank1}\textbf{23.4} & 7.8 & 9.0 & \cellcolor{rank2}23.9 & \cellcolor{rank1}\textbf{60.2} & 23.5 & 21.7 \\
\addlinespace[2pt]
 & Speedup & 1.00$\times$ & 1.87$\times$ & 3.93$\times$ & 3.76$\times$ & \cellcolor{abrank2}4.14$\times$ & \cellcolor{abrank1}\textbf{8.59$\times$} & 3.23$\times$ & 3.83$\times$ & \cellcolor{rank2}14.22$\times$ & \cellcolor{rank1}\textbf{30.08$\times$} & 13.72$\times$ & 12.82$\times$ \\
\addlinespace[3pt]
\rowcolor{gray!12}\multicolumn{14}{l}{\textbf{MMBench}} \\
\addlinespace[2pt]
 & Acc (\%) & \cellcolor{rank1}\textbf{81.9} & \cellcolor{rank2}79.3 & 63.9 & \cellcolor{abrank2}79.9 & 78.9$^{\dagger}$ & 70.3 & 78.5 & 79.3 & 77.8 & 69.0 & 78.1 & 78.1 \\
\addlinespace[2pt]
 & Latency (s) & 48.8 & 24.2 & 12.3 & \cellcolor{abrank2}11.0 & 11.9 & \cellcolor{abrank1}\textbf{5.7} & 12.1 & 13.4 & \cellcolor{rank2}5.4 & \cellcolor{rank1}\textbf{2.5} & 5.6 & 5.8 \\
\addlinespace[2pt]
 & TPS (tok/s)& 2.8 & 5.7 & 11.3 & 11.6 & \cellcolor{abrank2}11.7 & \cellcolor{abrank1}\textbf{25.3} & 11.5 & 10.4 & \cellcolor{rank2}23.6 & \cellcolor{rank1}\textbf{53.7} & 22.9 & 21.8 \\
\addlinespace[2pt]
 & Speedup & 1.00$\times$ & 2.02$\times$ & 3.97$\times$ & 2.22$\times$ & \cellcolor{abrank2}4.10$\times$ & \cellcolor{abrank1}\textbf{8.56$\times$} & 4.03$\times$ & 3.64$\times$ & \cellcolor{rank2}9.04$\times$ & \cellcolor{rank1}\textbf{19.52$\times$} & 8.71$\times$ & 8.41$\times$ \\
\addlinespace[3pt]
\rowcolor{gray!12}\multicolumn{14}{l}{\textbf{MME}} \\
\addlinespace[2pt]
 & Acc (\%) & 78.5 & 76.4 & 66.3 & 75.7 & \cellcolor{abrank1}\textbf{78.7} & 78.7 & \cellcolor{abrank1}\textbf{78.7} & \cellcolor{abrank1}\textbf{78.7} & \cellcolor{rank1}\textbf{78.9} & 78.7 & \cellcolor{rank1}\textbf{78.9} & \cellcolor{rank1}\textbf{78.9} \\
\addlinespace[2pt]
 & Latency (s) & 50.0 & 31.7 & 15.1 & \cellcolor{abrank2}9.9 & 11.6 & \cellcolor{abrank1}\textbf{7.7} & 13.9 & 13.8 & \cellcolor{rank2}5.5 & \cellcolor{rank1}\textbf{3.3} & 5.9 & 5.8 \\
\addlinespace[2pt]
 & TPS (tok/s)& 2.8 & 4.3 & 9.3 & \cellcolor{abrank2}12.8 & 12.1 & \cellcolor{abrank1}\textbf{21.4} & 10.6 & 10.7 & \cellcolor{rank2}23.9 & \cellcolor{rank1}\textbf{46.0} & 22.9 & 23.2 \\
\addlinespace[2pt]
 & Speedup & 1.00$\times$ & 1.58$\times$ & 3.31$\times$ & 2.53$\times$ & \cellcolor{abrank2}4.31$\times$ & \cellcolor{abrank1}\textbf{6.49$\times$} & 3.60$\times$ & 3.62$\times$ & \cellcolor{rank2}9.09$\times$ & \cellcolor{rank1}\textbf{15.15$\times$} & 8.47$\times$ & 8.62$\times$ \\
\addlinespace[3pt]
\rowcolor{gray!12}\multicolumn{14}{l}{\textbf{MMMU}} \\
\addlinespace[2pt]
 & Acc (\%) & \cellcolor{rank1}\textbf{42.9} & \cellcolor{rank2}42.4 & 37.2 & 42.5 & 41.6 & 41.0 & \cellcolor{abrank1}\textbf{43.0} & 41.8 & 41.6 & 41.6 & 41.5 & 41.5 \\
\addlinespace[2pt]
 & Latency (s) & 58.7 & 29.3 & 15.9 & \cellcolor{abrank2}10.5 & 15.0 & \cellcolor{abrank1}\textbf{8.9} & 17.9 & 15.5 & \cellcolor{rank2}7.5 & \cellcolor{rank1}\textbf{4.7} & 7.8 & 7.6 \\
\addlinespace[2pt]
 & TPS (tok/s)& 2.5 & 5.1 & 9.6 & \cellcolor{abrank2}12.1 & 10.3 & \cellcolor{abrank1}\textbf{21.0} & 9.0 & 10.0 & \cellcolor{rank2}20.8 & \cellcolor{rank1}\textbf{46.3} & 19.9 & 20.4 \\
\addlinespace[2pt]
 & Speedup & 1.00$\times$ & 2.00$\times$ & 3.69$\times$ & 2.80$\times$ & \cellcolor{abrank2}3.91$\times$ & \cellcolor{abrank1}\textbf{6.60$\times$} & 3.28$\times$ & 3.79$\times$ & \cellcolor{rank2}7.83$\times$ & \cellcolor{rank1}\textbf{12.49$\times$} & 7.53$\times$ & 7.72$\times$ \\
\addlinespace[3pt]
\rowcolor{gray!12}\multicolumn{14}{l}{\textbf{MM-Star}} \\
\addlinespace[2pt]
 & Acc (\%) & \cellcolor{rank1}\textbf{51.8} & \cellcolor{rank2}50.3 & 37.5 & 48.7 & \cellcolor{abrank2}51.8 & 51.7 & \cellcolor{abrank1}\textbf{51.9} & \cellcolor{abrank1}\textbf{51.9} & 47.6 & 47.6 & 47.5 & 47.5 \\
\addlinespace[2pt]
 & Latency (s) & 63.0 & 29.3 & 16.2 & \cellcolor{abrank2}10.1 & 15.3 & \cellcolor{abrank1}\textbf{9.6} & 17.4 & 17.0 & \cellcolor{rank2}5.8 & \cellcolor{rank1}\textbf{3.5} & 6.3 & 6.4 \\
\addlinespace[2pt]
 & TPS (tok/s)& 2.5 & 5.3 & 9.9 & \cellcolor{abrank2}12.6 & 11.3 & \cellcolor{abrank1}\textbf{25.2} & 10.2 & 10.4 & \cellcolor{rank2}23.5 & \cellcolor{rank1}\textbf{49.7} & 22.2 & 21.8 \\
\addlinespace[2pt]
 & Speedup & 1.00$\times$ & 2.15$\times$ & 3.89$\times$ & 3.12$\times$ & \cellcolor{abrank2}4.12$\times$ & \cellcolor{abrank1}\textbf{6.56$\times$} & 3.62$\times$ & 3.71$\times$ & \cellcolor{rank2}10.86$\times$ & \cellcolor{rank1}\textbf{18.00$\times$} & 10.00$\times$ & 9.84$\times$ \\
\addlinespace[3pt]
\rowcolor{gray!12}\multicolumn{14}{l}{\textbf{SEED-Bench}} \\
\addlinespace[2pt]
 & Acc (\%) & \cellcolor{rank1}\textbf{70.1}$^{\dagger}$ & 63.7$^{\dagger}$ & 39.3$^{\dagger}$ & 61.3$^{\dagger}$ & \cellcolor{abrank1}\textbf{70.4}$^{\dagger}$ & 70.3$^{\dagger}$ & \cellcolor{abrank1}\textbf{70.4}$^{\dagger}$ & \cellcolor{abrank1}\textbf{70.4}$^{\dagger}$ & 68.2$^{\dagger}$ & 68.2$^{\dagger}$ & \cellcolor{rank2}68.2$^{\dagger}$ & \cellcolor{rank2}68.2$^{\dagger}$ \\
\addlinespace[2pt]
 & Latency (s) & 50.3 & 25.4 & 12.5 & \cellcolor{abrank2}11.3 & 12.5 & \cellcolor{abrank1}\textbf{6.7} & 13.9 & 13.8 & \cellcolor{rank2}5.6 & \cellcolor{rank1}\textbf{3.0} & 5.7 & 5.8 \\
\addlinespace[2pt]
 & TPS (tok/s)& 2.7 & 5.3 & 10.8 & \cellcolor{abrank2}11.3 & 11.2 & \cellcolor{abrank1}\textbf{26.6} & 10.2 & 10.3 & \cellcolor{rank2}23.8 & \cellcolor{rank1}\textbf{55.5} & 23.3 & 23.1 \\
\addlinespace[2pt]
 & Speedup & 1.00$\times$ & 1.98$\times$ & 4.02$\times$ & 2.23$\times$ & \cellcolor{abrank2}4.02$\times$ & \cellcolor{abrank1}\textbf{7.51$\times$} & 3.62$\times$ & 3.64$\times$ & \cellcolor{rank2}8.98$\times$ & \cellcolor{rank1}\textbf{16.77$\times$} & 8.82$\times$ & 8.67$\times$ \\
\addlinespace[4pt]
\rowcolor{gray!12}\multicolumn{14}{l}{\textbf{Average}} \\
\addlinespace[2pt]
 & Acc (\%) & \cellcolor{rank1}\textbf{66.4} & 64.1 & 51.8 & 63.5 & 65.7 & 64.1 & \cellcolor{abrank1}\textbf{66.0} & \cellcolor{abrank2}65.8 & 64.5 & 63.0 & \cellcolor{rank2}64.5 & \cellcolor{rank2}64.5 \\
\addlinespace[2pt]
 & Latency (s) & 58.2 & 30.3 & 15.3 & \cellcolor{abrank2}10.5 & 14.2 & \cellcolor{abrank1}\textbf{8.0} & 16.6 & 15.7 & \cellcolor{rank2}5.9 & \cellcolor{rank1}\textbf{3.3} & 6.2 & 6.3 \\
\addlinespace[2pt]
 & TPS (tok/s)& 2.6 & 5.0 & 10.0 & \cellcolor{abrank2}12.1 & 11.1 & \cellcolor{abrank1}\textbf{23.8} & 9.9 & 10.1 & \cellcolor{rank2}23.3 & \cellcolor{rank1}\textbf{51.9} & 22.5 & 22.0 \\
\addlinespace[2pt]
 & Speedup & 1.00$\times$ & 1.93$\times$ & 3.80$\times$ & 2.78$\times$ & \cellcolor{abrank2}4.10$\times$ & \cellcolor{abrank1}\textbf{7.39$\times$} & 3.56$\times$ & 3.71$\times$ & \cellcolor{rank2}10.00$\times$ & \cellcolor{rank1}\textbf{18.67$\times$} & 9.54$\times$ & 9.35$\times$ \\
\bottomrule
\end{tabular}%
}
\vspace{-0.2cm}
\end{table*}

\subsection{Main Results}
\noindent\textbf{Accuracy-Speed Trade-Off.}
We report accuracy, wall-clock latency, throughput, and speedup in \cref{tab:performance} across all six benchmarks for three fixed-budget baselines, four full DRAMA variants, and five ablation configurations (detailed in \cref{sec:ablation}).
DRAMA CK~Prog and CK~Uni achieve ${\sim}9.5\times$ average speedup at $64.5\%$ accuracy ($-1.9$\,pp vs.\ the 64-step reference), simultaneously surpassing Baseline~32 in both speed (${\sim}5\times$ faster) and accuracy ($+0.4$\,pp).
FG~Prog offers a middle ground at $10.00\times$ and $64.5\%$ ($-1.9$\,pp), comparable to CK variants but with finer-grained stopping control.
At the aggressive end, FG~Uni pushes the speedup to $18.67\times$ at $63.0\%$ ($-3.4$\,pp).
All DRAMA variants substantially outperform Baseline~16 ($51.8\%$, $3.80\times$) in both quality and speed, confirming that analysis-guided acceleration yields strictly better Pareto trade-offs than uniform step reduction.
Notably, acceleration does not always come at an accuracy cost: on MME, every DRAMA variant surpasses the unaccelerated baseline ($78.9\%$ vs.\ $78.5\%$), and on AI2D, all variants match it, suggesting that pruning redundant visual tokens can alleviate attention dilution when the model's full visual budget exceeds the task's information requirement.
The accuracy-latency Pareto frontier (\cref{fig:pareto}) further illustrates this: DRAMA variants consistently dominate the front across all benchmarks, with no fixed-budget baseline achieving a comparable speed-accuracy combination.

\noindent\textbf{Throughput.}
In absolute throughput, the baseline's $2.6$\,tokens/s is impractical for interactive use. DRAMA CK variants raise this to $22.3$\,tokens/s, while FG~Uni reaches $51.9$\,tokens/s on average and peaks at $60.2$\,tokens/s on AI2D, bringing multimodal dLLM inference into a regime comparable to autoregressive models of similar scale.
Even the most conservative variant, CK~Prog, delivers a ${\sim}9\times$ throughput increase, confirming that the redundancy structures identified in \cref{sec:spatial} and \cref{sec:temporal} translate into substantial real-world speed gains.
\begin{figure}[t]
    \centering
    \includegraphics[width=\linewidth]{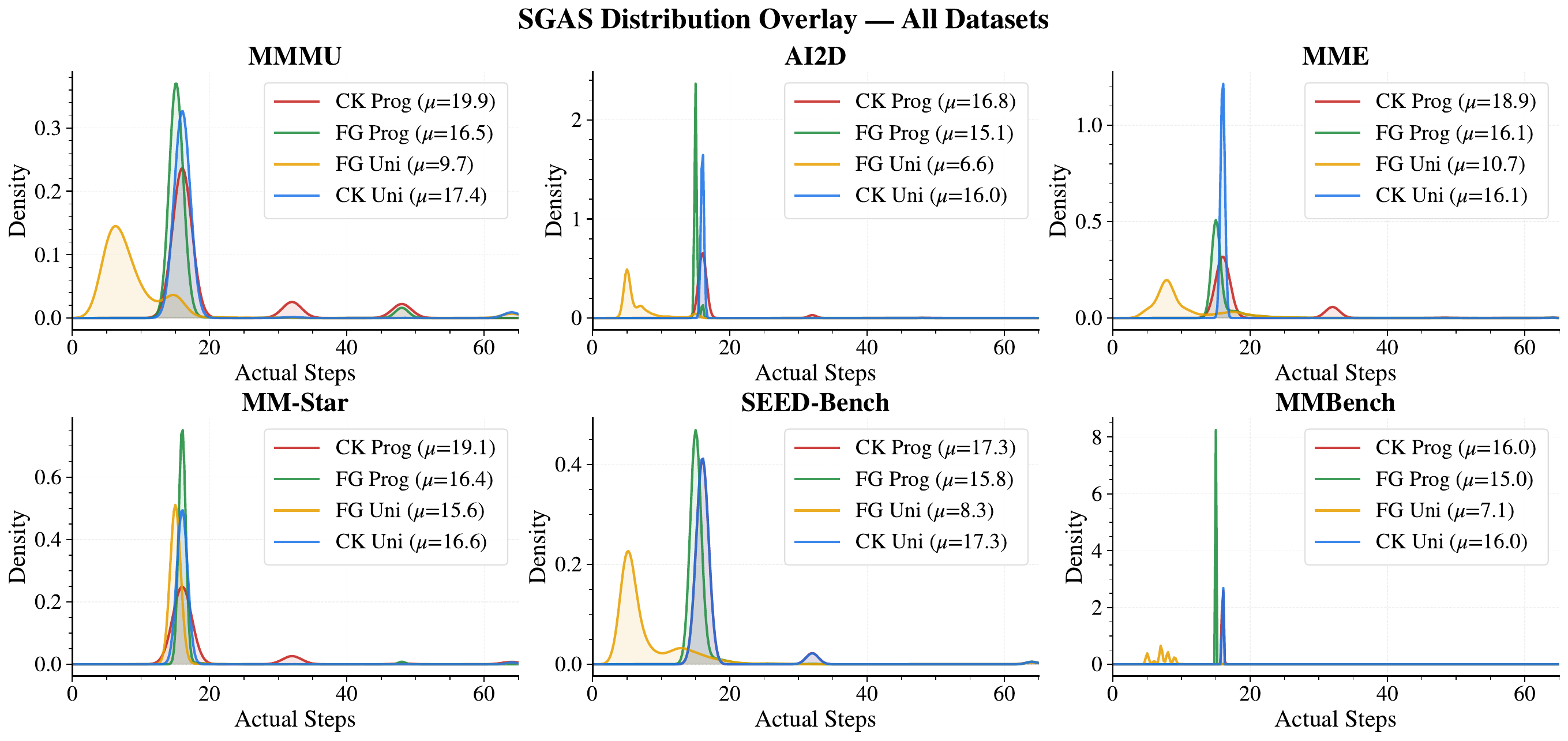}
    \vspace{-0.6cm}
    \caption{\textbf{Distribution of actual denoising steps} under SGAS across six benchmarks. Each subplot shows the histogram and KDE for one SGAS variant. FG~Uni concentrates most samples at very early steps ($\mu\!\approx\!8$--$12$), while CK variants cluster around the first eligible checkpoint (step~16). Progressive scheduling shifts the distribution rightward, reflecting its stricter early-step requirements.}
    \vspace{-0.5cm}
    \label{fig:step_dist}
\end{figure}

\noindent\textbf{Per-Benchmark Variation.}
Acceleration gains vary with task characteristics.
On AI2D, FG~Uni achieves $30.08\times$ speedup while matching the 64-step baseline accuracy ($73.0\%$), as diagram-centric queries concentrate attention on a few salient patches.
On MME, CK variants simultaneously improve accuracy ($78.9\%$ vs.\ $78.5\%$) and deliver up to $8.47\times$ speedup, reflecting rapid attention convergence for perceptual tasks.
On MMMU, CK variants retain $41.5\%$ ($-1.4$\,pp) at ${\sim}7.5\times$ speedup; on MM-Star, $47.5\%$ ($-4.3$\,pp) at ${\sim}10\times$; MMBench follows a similar pattern at $78.1\%$ ($-3.8$\,pp, ${\sim}8.5\times$). We further disentangle the contribution of each component in \cref{sec:ablation}, where SGAS alone matches accuracy on most of the benchmarks, confirming that it serves as a stable backbone that absorbs the accuracy cost of aggressive visual token pruning.

\subsection{Ablation Study}
\label{sec:ablation}
We further ablate (i) the individual contributions of SVTP and SGAS as orthogonal cost-axis reductions; and (ii) the interaction between SGAS's evaluation-granularity and threshold-scheduling axes that defines the 2$\times$2 design space introduced in \cref{sec:stopping}. We defer (iii) generalisation to a second backbone (LaViDa); and (iv) head-to-head comparisons with two concurrent multimodal dLLM acceleration methods to Appendix~\ref{app:ablation}.

\noindent\textbf{Effect of SVTP.}
At a fixed budget of $T\!=\!32$ steps, SVTP alone yields $2.78\times$ normalised speedup at $63.5\%$ accuracy, reducing per-sample time from $30.3$\,s to $10.5$\,s (see the ablation columns of \cref{tab:performance}).
The gain comes entirely from shortening the key-value sequence, consistent with the spatial concentration in \cref{sec:spatial}, and accuracy remains within $-2.4$ to $+0.6$\,pp of the 32 steps baseline on all benchmarks.
When combined with SGAS, SVTP contributes a consistent ${\sim}2.5\times$ multiplicative factor regardless of the stopping variant.
This confirms that spatial pruning and temporal stopping target orthogonal cost axes ($N$ and $T$ in $\mathcal{O}(T \cdot L \cdot N^2)$) and compose independently.

\noindent\textbf{Effect of SGAS.}
Without SVTP, all four SGAS configurations achieve $3.56$--$7.39\times$ speedup while staying within $2.3$\,pp of the 64 steps baseline on average (see the ablation columns of \cref{tab:performance}).
The most conservative variant, CK~Prog, matches or exceeds the 64-step baseline's accuracy on five of six benchmarks at $3.56\times$ speedup, indicating that the fixed budget over-denoises on most tasks.
At the aggressive end, FG~Uni reaches $7.39\times$ while matching Baseline~32 accuracy.
This confirms that the temporal redundancy identified in \cref{sec:temporal} translates directly into our SGAS component without dedicated retraining.

\noindent\textbf{Stopping Strategy Interaction.}
We discover that evaluation granularity (FG vs.\ CK) and threshold schedule (Prog vs.\ Uni) interact strongly.
Under uniform thresholds, FG nearly doubles CK's speedup, aligning with per-step evaluation's more exit opportunities.
Under progressive thresholds, this gap largely vanishes: the stricter early-step requirements absorb FG's frequent evaluations, keeping accuracy close to the baseline.
Symmetrically, switching from Prog to Uni barely affects CK ($3.56\times vs. 3.71\times$) but dramatically shifts FG ($4.1\times vs. 7.4\times$).
We visualize DRAMA's stopping-step distributions (\cref{fig:step_dist}, per-benchmark breakdowns are shown in Supplementary Materials), which corroborate our JSD redundancy analysis: FG~Uni concentrates exits at early steps, CK variants cluster around the first checkpoint, and progressive scheduling shifts the distribution rightward.
This shows that progressive scheduling acts as a safeguard during the structure-formation phase (\cref{sec:temporal}), while still allowing the system to exploit plateau-phase redundancy for substantial acceleration.

\section{Conclusion}
We present DRAMA, a training-free inference acceleration framework for multimodal dLLMs that closes the gap between merely observing attention redundancy and systematically exploiting it. Driven by a measured dissection of where redundancy lies, DRAMA delivers an average 18.67$\times$ inference speedup on LLaDA-V across six vision-language benchmarks with competitive accuracy, attained without any training, architectural modification, or attention caching. Beyond its empirical gains, DRAMA reframes acceleration in this emerging paradigm as a problem of identifying and exploiting structural redundancy, offering an analysis-driven perspective which informs broader efforts toward measurement-grounded acceleration of multimodal diffusion large language models.

\begin{ack}
This work was supported by the National Science and Technology Council (NSTC) of Taiwan under Grant 115-2634-F-002-012 and by the NTU AI Center of Research Excellence (AI-CoRE), with additional computational support provided through H100 GPUs donated by Wistron Corporation. Shao-Yuan Lo is honored to be supported by the Yushan Young Fellow Program by the Ministry of Education (MOE) of Taiwan.
\end{ack}

\medskip
{
\small
\bibliographystyle{plainnat}
\bibliography{main}
}

\newpage
\appendix
\section{Limitations and Potential Negative Social Impact}
\subsection{Limitations}
\label{sec:appendix_limitations}
Our work accelerates inference for multimodal dLLMs without modifying the underlying model weights or training data. As such, it is not intended to introduce new capabilities beyond those already present in the base model. Rather than improving the model’s intrinsic reasoning or multimodal understanding ability, our method focuses on making existing capabilities more efficient at inference time. This limitation is shared by prior training-free acceleration methods~\cite{ma2025dkv, liu2025dllm, wu2025fast1, nguyen2025attention, xu2025redvtp, chang2026d3tom}, whose primary goal is to reduce latency and computation while preserving model quality as much as possible, instead of advancing the underlying performance frontier.

\subsection{Potential Negative Social Impact}
By substantially lowering the computational cost of deploying multimodal dLLMs, DRAMA may also reduce the barrier for malicious use, including the large-scale generation of misleading or deceptive visual question answering outputs~\cite{liu2024survey, bai2024hallucination}. Such misuse could amplify the spread of incorrect visual interpretations or plausible-sounding but unreliable multimodal responses. This concern is not specific to our method, but is shared broadly by inference acceleration techniques that improve the accessibility and scalability of foundation models.
\section{Additional Implementation Details}
\label{sec:appendix_impl}
\subsection{Datasets and Evaluation Splits}
\label{sec:appendix_splits}
We evaluate on six standard vision-language benchmarks. For AI2D~\cite{kembhavi2016diagram}, MMBench~\cite{liu2024mmbench}, MME~\cite{fu2023mme}, MMMU~\cite{yue2024mmmu}, and MM-Star~\cite{chen2024we}, we use the full evaluation split released by each benchmark. For SEED-Bench~\cite{li2023seed}, we evaluate on a fixed random subset of 2{,}000 samples (seed fixed for reproducibility) to manage computational cost; the same subset is used across all baselines, ablations, and DRAMA variants to ensure fair comparison. \Cref{tab:datasets} summarizes the splits and sample counts.
\begin{table}[h]
\centering
\caption{Evaluation splits used in our experiments.}
\label{tab:datasets}
\footnotesize
\begin{tabular}{llc}
\toprule
\textbf{Benchmark} & \textbf{Split} & \textbf{\#Samples} \\
\midrule
AI2D~\cite{kembhavi2016diagram}        & test (full)              & 3{,}088 \\
MMBench~\cite{liu2024mmbench}          & dev-en (full)            & 4{,}329 \\
MME~\cite{fu2023mme}                   & full                     & 2{,}374 \\
MMMU~\cite{yue2024mmmu}                & validation (full)        & 1{,}050 \\
MM-Star~\cite{chen2024we}              & full                     & 1{,}500 \\
SEED-Bench~\cite{li2023seed}           & random subset (seed-fixed) & 2{,}000 \\
\bottomrule
\end{tabular}
\end{table}

\subsection{Prompt Template}
\label{sec:appendix_prompt}

Following LLaDA-V~\cite{you2025llada}, we use a multi-turn chat format with a system message (``\textit{You are a helpful language and vision assistant\ldots}''), a user turn containing the image token, question, and a chain-of-thought thinking instruction, and an empty assistant turn for the model to complete. All experiments use a fixed generation length of $N_g\!=\!128$ tokens with deterministic decoding (temperature $0.0$) and low-confidence remasking. \Cref{tab:prompt} shows an example of a prompt template used for our evaluation benchmarks.

\begin{table}[h]
\centering
\caption{\textbf{An example of prompt template}}
\label{tab:prompt}
\resizebox{\linewidth}{!}{%
\begin{tabular}{p{2.8cm} p{10.5cm}}
\toprule
\textbf{Component} & \textbf{Content} \\
\midrule
Image token & \texttt{<image>} \\
\addlinespace[4pt]
Question & \texttt{Question: \{question\}} \\
\addlinespace[4pt]
Options & \texttt{Options:} \newline \texttt{A. \{option\_a\}} \newline \texttt{B. \{option\_b\}} \newline \texttt{...} \\
\addlinespace[4pt]
Thinking \newline instruction & \textit{You should first think about the reasoning process in the mind and then provide the user with the answer. The reasoning process is enclosed within} \texttt{<think>} \texttt{</think>} \textit{tags, i.e.}\ \texttt{<think>} \textit{reasoning process here} \texttt{</think>} \texttt{<answer>} \textit{answer here:} Please select the correct option by responding with only the letter (A, B, C, D, or E) \texttt{</answer>} \\
\bottomrule
\end{tabular}%
}
\end{table}
\subsection{SGAS Configuration Details}
\label{sec:appendix_sgas}
\begin{table}[h]
\centering
\caption{SGAS configuration variants. All variants share $T\!=\!64$, $T_{\min}\!=\!1$.}
\label{tab:sgas_config}
\footnotesize
\begin{tabular}{lccl}
\toprule
\textbf{Variant} & \textbf{Granularity} & \textbf{Schedule} & \textbf{Evaluation Steps} \\
\midrule
CK Prog & Checkpoint & Progressive & $\mathcal{C} = \{16, 32, 48, 64\}$ \\
CK Uni  & Checkpoint & Uniform     & $\mathcal{C} = \{16, 32, 48, 64\}$ \\
FG Prog & Fine-grained & Progressive & Every step $t > T_{\min}$ \\
FG Uni  & Fine-grained & Uniform     & Every step $t > T_{\min}$ \\
\bottomrule
\end{tabular}
\end{table}
We summarize the four SGAS variants evaluated in the main paper in~\cref{tab:sgas_config}. Under progressive scheduling, the stopping criterion is strict at early evaluation steps, requiring both semantic sufficiency and high lexical coherence, and progressively relaxed at later steps, reflecting the diminishing marginal return of additional denoising iterations identified in the temporal analysis of the main paper.
Under uniform scheduling, fixed thresholds are applied throughout all evaluation steps.
Checkpoint mode only evaluates at the sparse set $\mathcal{C}$, while fine-grained mode evaluates at every step beyond $T_{\min}$, providing more exit opportunities at the cost of slightly more frequent criterion checks.
\subsubsection{Stopping Decision Procedure}
We formalize the full SGAS procedure in~\cref{alg:stopping}. After each denoising step, if $t \geq T_{\min}$ and $t$ is admitted by the granularity schedule $\mathcal{E}$, the decoded sequence is evaluated against two complementary checks. The \emph{semantic sufficiency} check parses the decoded text under the task schema. The \emph{lexical coherence} check verifies three statistics of $\mathbf{x}_t$ against the threshold vector $\boldsymbol{\theta}_t = (\theta^{\mathrm{div}}_t, \theta^{\mathrm{rep}}_t, \theta^{\mathrm{con}}_t)$:
\begin{itemize}
    \item Vocabulary diversity $r^{\mathrm{div}}$: ratio of unique tokens to total non-mask tokens.
    \item Max single-token frequency $r^{\mathrm{rep}}$: fraction of the most repeated token.
    \item Consecutive-repetition ratio $r^{\mathrm{con}}$: fraction of adjacent token pairs that are identical.
\end{itemize}
The coherence gate passes when $r^{\mathrm{div}} \!\geq\! \theta^{\mathrm{div}}_t$, $r^{\mathrm{rep}} \!\leq\! \theta^{\mathrm{rep}}_t$, and $r^{\mathrm{con}} \!\leq\! \theta^{\mathrm{con}}_t$ simultaneously. If both checks pass, denoising terminates and $\mathbf{x}_t$ is returned; otherwise the procedure proceeds to step $t{+}1$. If no step in the trajectory triggers the predicate, $\mathbf{x}_T$ is returned, recovering the unaccelerated baseline as a worst case.

\begin{algorithm}[h]
\caption{Sufficiency-Guided Adaptive Stopping}
\label{alg:stopping}
\begin{algorithmic}[1]
\Require Denoising budget $T$, minimum steps $T_{\min}$, evaluation set $\mathcal{E}$ \Comment{$\mathcal{C}$ or $\{T_{\min}\!+\!1,\ldots,T\}$}
\Require Threshold schedule $\theta(\cdot) = (\theta^{\mathrm{div}}, \theta^{\mathrm{rep}}, \theta^{\mathrm{con}})$ \Comment{progressive or uniform}
\State $\mathbf{x}_0 \leftarrow [\textsc{mask}]^{N_g}$
\For{$t = 1, \ldots, T$}
    \State $\mathbf{x}_t \leftarrow \textsc{Denoise}(\mathbf{x}_{t-1})$ \Comment{Predict \& unmask ${\sim} N_g/T$ tokens}
    \If{$t < T_{\min}$ \textbf{or} $t \notin \mathcal{E}$}
        \State \textbf{continue}
    \EndIf
    \State $(\theta^{\mathrm{div}}_t, \theta^{\mathrm{rep}}_t, \theta^{\mathrm{con}}_t) \leftarrow \theta(t)$
    \State $\textit{text} \leftarrow \textsc{Decode}(\mathbf{x}_t)$
    \State $\textit{sufficient} \leftarrow \textsc{SemanticSufficiencyCheck}(\textit{text}, \textit{task})$
    \State $r^{\mathrm{div}} \leftarrow \textsc{VocabularyDiversity}(\mathbf{x}_t)$
    \State $r^{\mathrm{rep}} \leftarrow \textsc{MaxTokenFrequency}(\mathbf{x}_t)$
    \State $r^{\mathrm{con}} \leftarrow \textsc{ConsecutiveRepetitionRatio}(\mathbf{x}_t)$
    \State $\textit{coherent} \leftarrow (r^{\mathrm{div}} \!\geq\! \theta^{\mathrm{div}}_t) \wedge (r^{\mathrm{rep}} \!\leq\! \theta^{\mathrm{rep}}_t) \wedge (r^{\mathrm{con}} \!\leq\! \theta^{\mathrm{con}}_t)$
    \If{$\textit{sufficient}$ \textbf{and} $\textit{coherent}$}
        \State \Return $\mathbf{x}_t$ \Comment{Early termination}
    \EndIf
\EndFor
\State \Return $\mathbf{x}_T$ \Comment{Fallback: full budget}
\end{algorithmic}
\end{algorithm}

\subsubsection{Threshold Parameters}
\Cref{tab:sgas_thresholds} lists the threshold vectors used by each variant. Checkpoint variants use a set of per-checkpoint thresholds; fine-grained uniform uses a single fixed vector; and fine-grained progressive smoothly interpolates thresholds as a function of denoising progress $p = t/T$.
\begin{table}[h]
\centering
\caption{Lexical coherence thresholds for all SGAS variants ($T\!=\!64$).}
\label{tab:sgas_thresholds}
\footnotesize
\begin{tabular}{llccc}
\toprule
\textbf{Variant} & \textbf{Step / Phase} & $\theta^{\text{div}}$ $\uparrow$ & $\theta^{\text{rep}}$ $\downarrow$ & $\theta^{\text{con}}$ $\downarrow$ \\
\midrule
\multirow{4}{*}{CK Uni}
& $t = 16$ & 0.50 & 0.22 & 0.09 \\
& $t = 32$ & 0.50 & 0.22 & 0.09 \\
& $t = 48$ & 0.50 & 0.22 & 0.09 \\
& $t = 64$ & 0.50 & 0.22 & 0.09 \\
\midrule
\multirow{4}{*}{CK Prog}
& $t = 16$ & 0.50 & 0.23 & 0.09 \\
& $t = 32$ & 0.55 & 0.20 & 0.08 \\
& $t = 48$ & 0.50 & 0.23 & 0.09 \\
& $t = 64$ & 0.45 & 0.25 & 0.10 \\
\midrule
FG Uni & All steps & 0.50 & 0.22 & 0.09 \\
\midrule
\multirow{2}{*}{FG Prog}
& $p \leq 0.5$ (tighten) & $0.45 \!\to\! 0.55$ & $0.25 \!\to\! 0.20$ & $0.10 \!\to\! 0.08$ \\
& $p > 0.5$ (relax) & $0.55 \!\to\! 0.45$ & $0.20 \!\to\! 0.25$ & $0.08 \!\to\! 0.10$ \\
\bottomrule
\end{tabular}
\end{table}

\subsection{Compatibility with Block Diffusion}
\label{sec:appendix_block}
Block diffusion~\cite{arriola2025block} is a semi-autoregressive generation paradigm for dLLMs in which the $N_g$ generation tokens are partitioned into $B$ blocks of size $b\!=\!N_g/B$, with each block denoised sequentially over its own trajectory of $T_b$ steps; it is a generation strategy rather than an acceleration mechanism. Our experiments adopt the default single-block setting ($B\!=\!1$) of LLaDA-V~\cite{you2025llada}, but DRAMA extends to the multi-block regime without modification. The visual prefix is shared across all blocks, so SVTP performs its cross-modal scoring once before the first block and reuses the selected patch set $\mathcal{P}$ for every subsequent block. SGAS evaluates its stop predicate inside each block independently, using the same threshold schedule $\boldsymbol{\theta}(t)$, so each block terminates at its own first-hit step $t^{\star}_b$ without coupling across blocks. The aggregate skipped-step count is then $\sum_{b=1}^{B}(T_b\!-\!t^{\star}_b)$.

\subsection{Hyperparameter selection.}
\label{sec:appendix_hyper}
All settings reported in this paper are fixed once and shared across every benchmark and both backbones; we perform no per-benchmark tuning and no search. SVTP retains a single patch, $|\mathcal{P}| = 1$, in all experiments. The SGAS threshold vectors of Table~\ref{tab:sgas_thresholds} were set from the output statistics of 100 LLaDA-V samples on MMMU, and the resulting values were then applied unchanged to the remaining five benchmarks and to LaViDa. Accuracy on the benchmarks we report was therefore never consulted when choosing these values.
\section{Detailed Ablation Study}
\label{app:ablation}
\subsection{Model Architecture Ablation}
We further evaluate DRAMA on LaViDa~\cite{li2025lavida}, a multimodal dLLM that already employs $2{\times}2$ visual-token pooling and therefore exposes a substantially smaller acceleration headroom than LLaDA-V (LaViDa baseline averages $9.76$\,s vs.\ LLaDA-V's $58.2$\,s under the same think-mode regime). All LaViDa results are evaluated on the full (AI2D $3088$; MME $2374$; MM-Star $1500$; MMMU $900$) benchmark and a 1,000-sample SEED-Bench subset . Despite the compressed margin, DRAMA delivers consistent $1.73$--$1.84\times$ wall-clock speedup on average across five benchmarks (\cref{tab:performance_lavida}), with accuracy preserved to $97.2\%$ overall by CK~Uni and $97.0\%$ by FG~Uni, and with MME improving over the LaViDa baseline of 68.91, reaching 69.46 under SVTP alone and 69.17 under DRAMA FG~Uni and CK~Uni, indicating that pruning redundant visual tokens can mitigate attention dilution even when the backbone has already partially compressed its visual context. The pattern of variant trade-offs largely mirrors LLaDA-V: FG~Uni and FG~Prog dominate the speedup axis, while CK~Uni retains the most accuracy, so the original split between the two schedules still holds under full-set evaluation.

\begin{table*}[t]
\centering
\caption{Benchmark results on LaViDa (128 denoising steps, think mode). Sample counts are given in the benchmark headers. Latency is wall-clock seconds per sample and TPS is computed from the fixed generation length of 128 tokens under the measured latency. Metrics: Acc (\%, $\uparrow$), Latency (s/sample, $\downarrow$), TPS (tokens/s, $\uparrow$), Speedup vs.\ the 128-step baseline ($\uparrow$), Acc Retention ($\uparrow$).}
\label{tab:performance_lavida}
\resizebox{\textwidth}{!}{%
\begin{tabular}{l l c c c c c c c c c c}
\toprule
\multicolumn{2}{c}{} & \multicolumn{1}{c}{\textbf{Baseline}} & \multicolumn{5}{c}{\textcolor{gray}{\textbf{Ablation}}} & \multicolumn{4}{c}{\textbf{DRAMA (Ours)}} \\
\cmidrule(lr){3-3} \cmidrule(lr){4-8} \cmidrule(lr){9-12}
 & \textbf{Metric}
& \makecell[c]{\footnotesize 128 \\ \footnotesize steps}
& \makecell[c]{\textcolor{gray}{\footnotesize w/o} \\ \textcolor{gray}{\footnotesize SGAS}}
& \makecell[c]{\textcolor{gray}{\footnotesize w/o SVTP} \\ \textcolor{lightgray}{\footnotesize FG Prog}}
& \makecell[c]{\textcolor{gray}{\footnotesize w/o SVTP} \\ \textcolor{lightgray}{\footnotesize FG Uni}}
& \makecell[c]{\textcolor{gray}{\footnotesize w/o SVTP} \\ \textcolor{lightgray}{\footnotesize CK Prog}}
& \makecell[c]{\textcolor{gray}{\footnotesize w/o SVTP} \\ \textcolor{lightgray}{\footnotesize CK Uni}}
& \makecell[c]{\footnotesize FG Prog}
& \makecell[c]{\footnotesize FG Uni}
& \makecell[c]{\footnotesize CK Prog}
& \makecell[c]{\footnotesize CK Uni} \\
\midrule
\addlinespace[3pt]
\rowcolor{gray!12}\multicolumn{12}{l}{\textbf{AI2D} \footnotesize ($n=3088$)} \\
\addlinespace[2pt]
 & Acc (\%) & \cellcolor{abrank1}\textbf{69.88} & \cellcolor{abrank2}69.04 & 68.36 & 68.72 & 68.49 & 68.75 & 67.58 & 67.81 & 67.81 & 67.45 \\
\addlinespace[2pt]
 & Latency (s) & 9.98 & 7.40 & 7.04 & 7.07 & 7.49 & 7.50 & \cellcolor{abrank2}5.30 & \cellcolor{abrank1}\textbf{5.23} & 5.50 & 5.54 \\
\addlinespace[2pt]
 & TPS (tok/s) & 12.82 & 17.29 & 18.17 & 18.11 & 17.08 & 17.06 & \cellcolor{abrank2}24.13 & \cellcolor{abrank1}\textbf{24.49} & 23.27 & 23.11 \\
\addlinespace[2pt]
 & Speedup & 1.00$\times$ & 1.35$\times$ & 1.42$\times$ & 1.41$\times$ & 1.33$\times$ & 1.33$\times$ & \cellcolor{abrank2}1.88$\times$ & \cellcolor{abrank1}\textbf{1.91$\times$} & 1.82$\times$ & 1.80$\times$ \\
\addlinespace[2pt]
 & Acc Retention & \cellcolor{abrank1}\textbf{1.00} & \cellcolor{abrank2}0.99 & 0.98 & 0.98 & 0.98 & 0.98 & 0.97 & 0.97 & 0.97 & 0.97 \\
\addlinespace[4pt]
\addlinespace[3pt]
\rowcolor{gray!12}\multicolumn{12}{l}{\textbf{MME} \footnotesize ($n=2374$)} \\
\addlinespace[2pt]
 & Acc (\%) & 68.91 & \cellcolor{abrank1}\textbf{69.46} & 68.32 & 68.62 & 66.51 & 68.62 & 68.53 & \cellcolor{abrank2}69.17 & 62.22 & 69.17 \\
\addlinespace[2pt]
 & Latency (s) & 9.82 & 7.15 & 7.06 & 7.24 & 7.04 & 7.72 & \cellcolor{abrank1}\textbf{5.34} & 5.40 & \cellcolor{abrank2}5.39 & 5.70 \\
\addlinespace[2pt]
 & TPS (tok/s) & 13.04 & 17.89 & 18.12 & 17.67 & 18.19 & 16.58 & \cellcolor{abrank1}\textbf{23.97} & 23.72 & \cellcolor{abrank2}23.76 & 22.44 \\
\addlinespace[2pt]
 & Speedup & 1.00$\times$ & 1.37$\times$ & 1.39$\times$ & 1.35$\times$ & 1.40$\times$ & 1.27$\times$ & \cellcolor{abrank1}\textbf{1.84$\times$} & 1.82$\times$ & \cellcolor{abrank2}1.82$\times$ & 1.72$\times$ \\
\addlinespace[2pt]
 & Acc Retention & 1.00 & \cellcolor{abrank1}\textbf{1.01} & 0.99 & 1.00 & 0.97 & 1.00 & 0.99 & \cellcolor{abrank2}1.00 & 0.90 & 1.00 \\
\addlinespace[4pt]
\addlinespace[3pt]
\rowcolor{gray!12}\multicolumn{12}{l}{\textbf{MMMU} \footnotesize ($n=900$)} \\
\addlinespace[2pt]
 & Acc (\%) & \cellcolor{abrank1}\textbf{42.44} & 41.78 & 41.67 & 42.11 & 42.00 & \cellcolor{abrank1}\textbf{42.44} & 41.11 & 41.22 & 41.44 & 42.00 \\
\addlinespace[2pt]
 & Latency (s) & 9.91 & 7.68 & 7.06 & 7.19 & 7.47 & 7.60 & \cellcolor{abrank1}\textbf{5.52} & \cellcolor{abrank2}5.65 & 5.77 & 5.84 \\
\addlinespace[2pt]
 & TPS (tok/s) & 12.91 & 16.66 & 18.13 & 17.79 & 17.14 & 16.84 & \cellcolor{abrank1}\textbf{23.19} & \cellcolor{abrank2}22.66 & 22.18 & 21.90 \\
\addlinespace[2pt]
 & Speedup & 1.00$\times$ & 1.29$\times$ & 1.40$\times$ & 1.38$\times$ & 1.33$\times$ & 1.30$\times$ & \cellcolor{abrank1}\textbf{1.80$\times$} & \cellcolor{abrank2}1.75$\times$ & 1.72$\times$ & 1.70$\times$ \\
\addlinespace[2pt]
 & Acc Retention & \cellcolor{abrank1}\textbf{1.00} & 0.98 & 0.98 & 0.99 & 0.99 & \cellcolor{abrank1}\textbf{1.00} & 0.97 & 0.97 & 0.98 & 0.99 \\
\addlinespace[3pt]
\rowcolor{gray!12}\multicolumn{12}{l}{\textbf{MM-Star} \footnotesize ($n=1500$)} \\
\addlinespace[2pt]
 & Acc (\%) & \cellcolor{abrank1}\textbf{43.00} & 40.73 & 41.33 & \cellcolor{abrank2}42.47 & 41.33 & 42.20 & 39.13 & 40.00 & 39.00 & 40.00 \\
\addlinespace[2pt]
 & Latency (s) & 9.19 & 7.31 & 6.50 & 6.58 & 6.87 & 6.99 & \cellcolor{abrank1}\textbf{5.23} & \cellcolor{abrank2}5.36 & 5.63 & 5.57 \\
\addlinespace[2pt]
 & TPS (tok/s) & 13.93 & 17.52 & 19.68 & 19.45 & 18.64 & 18.32 & \cellcolor{abrank1}\textbf{24.46} & \cellcolor{abrank2}23.89 & 22.74 & 22.97 \\
\addlinespace[2pt]
 & Speedup & 1.00$\times$ & 1.26$\times$ & 1.41$\times$ & 1.40$\times$ & 1.34$\times$ & 1.32$\times$ & \cellcolor{abrank1}\textbf{1.76$\times$} & \cellcolor{abrank2}1.72$\times$ & 1.63$\times$ & 1.65$\times$ \\
\addlinespace[2pt]
 & Acc Retention & \cellcolor{abrank1}\textbf{1.00} & 0.95 & 0.96 & \cellcolor{abrank2}0.99 & 0.96 & 0.98 & 0.91 & 0.93 & 0.91 & 0.93 \\
\addlinespace[4pt]
\addlinespace[3pt]
\rowcolor{gray!12}\multicolumn{12}{l}{\textbf{SEED-Bench} \footnotesize ($n=1000$)} \\
\addlinespace[2pt]
 & Acc (\%) & \cellcolor{abrank1}\textbf{63.50} & 61.00 & 61.40 & \cellcolor{abrank2}63.40 & 61.40 & 63.40 & 59.10 & 61.00 & 59.10 & 61.00 \\
\addlinespace[2pt]
 & Latency (s) & 9.90 & 7.34 & 7.09 & 7.21 & 7.52 & 7.67 & \cellcolor{abrank1}\textbf{5.19} & \cellcolor{abrank2}5.24 & 5.55 & 5.61 \\
\addlinespace[2pt]
 & TPS (tok/s) & 12.93 & 17.45 & 18.06 & 17.75 & 17.02 & 16.68 & \cellcolor{abrank1}\textbf{24.67} & \cellcolor{abrank2}24.45 & 23.06 & 22.82 \\
\addlinespace[2pt]
 & Speedup & 1.00$\times$ & 1.35$\times$ & 1.40$\times$ & 1.37$\times$ & 1.32$\times$ & 1.29$\times$ & \cellcolor{abrank1}\textbf{1.91$\times$} & \cellcolor{abrank2}1.89$\times$ & 1.78$\times$ & 1.77$\times$ \\
\addlinespace[2pt]
 & Acc Retention & \cellcolor{abrank1}\textbf{1.00} & 0.96 & 0.97 & \cellcolor{abrank2}1.00 & 0.97 & 1.00 & 0.93 & 0.96 & 0.93 & 0.96 \\
\addlinespace[4pt]
\rowcolor{gray!12}\multicolumn{12}{l}{\textbf{Average} \footnotesize (over the five benchmarks above)} \\
\addlinespace[2pt]
 & Acc (\%) & \cellcolor{abrank1}\textbf{57.55} & 56.40 & 56.22 & 57.06 & 55.95 & \cellcolor{abrank2}57.08 & 55.09 & 55.84 & 53.91 & 55.92 \\
\addlinespace[2pt]
 & Latency (s) & 9.76 & 7.38 & 6.95 & 7.06 & 7.28 & 7.50 & \cellcolor{abrank1}\textbf{5.32} & \cellcolor{abrank2}5.38 & 5.57 & 5.65 \\
\addlinespace[2pt]
 & TPS (tok/s) & 13.11 & 17.35 & 18.42 & 18.14 & 17.59 & 17.08 & \cellcolor{abrank1}\textbf{24.08} & \cellcolor{abrank2}23.81 & 22.99 & 22.65 \\
\addlinespace[2pt]
 & Speedup & 1.00$\times$ & 1.32$\times$ & 1.40$\times$ & 1.38$\times$ & 1.34$\times$ & 1.30$\times$ & \cellcolor{abrank1}\textbf{1.84$\times$} & \cellcolor{abrank2}1.82$\times$ & 1.75$\times$ & 1.73$\times$ \\
\addlinespace[2pt]
 & Acc Retention & \cellcolor{abrank1}\textbf{1.00} & 0.98 & 0.98 & \cellcolor{abrank2}0.99 & 0.97 & 0.99 & 0.96 & 0.97 & 0.94 & 0.97 \\
\bottomrule
\end{tabular}%
}
\end{table*}

\subsection{Comparison with Concurrent Acceleration Methods}
\subsubsection{Comparison with RedVTP~\cite{xu2025redvtp}}
RedVTP~\cite{xu2025redvtp} accelerates multimodal dLLMs by pruning visual tokens, and is the closest prior method on our backbone. We re-run it ourselves at its most aggressive operating point ($r{=}0.25$) under our own 32-step think-mode setting, against a baseline with the same batch and the same 500 samples.
Table~\ref{tab:redvtp_lladav} pairs each method with its own baseline on LLaDA-V. RedVTP buys $1.21$--$1.28\times$ while retaining $84.8$--$98.1\%$ of baseline accuracy, averaging $1.25\times$ at $91.5\%$. DRAMA reaches $8.32$--$16.32\times$ at $94.7$--$100.4\%$ retention, averaging $12.32\times$ at $97.6\%$. Pruning the visual prefix removes work that is paid once per sample, so its gain is bounded by the share of the forward pass that the prefix occupies; the step axis is not bounded in the same way, and the two axes compose.
The same ordering holds on LaViDa (Table~\ref{tab:redvtp_lavida}), where RedVTP reports only $r{=}75\%$ because the $2{\times}2$ visual-token pooling already built into LaViDa absorbs most of its pruning headroom. Its wall-clock gain there is a marginal $1.06$--$1.08\times$, against $1.56$--$1.91\times$ for DRAMA under a heavier reasoning regime. That the ordering survives a backbone whose visual prefix is already compressed supports the reading that the temporal redundancy identified in Section~\ref{sec:temporal} is a separate, multiplicative source of acceleration rather than a restatement of spatial redundancy.

\begin{table}[t]
\renewcommand{\arraystretch}{1.1}
\centering
\caption{Per-benchmark comparison with RedVTP on LLaDA-V, both reproduced by us. RedVTP runs at its most aggressive operating point ($r{=}0.25$) under the same 32-step think-mode regime as DRAMA.}
\label{tab:redvtp_lladav}
\resizebox{\columnwidth}{!}{%
\begin{tabular}{l l l c c c c}
\toprule
\textbf{Benchmark} & \textbf{Method} & \textbf{Setting} & \textbf{Performance} & \textbf{Latency (s)} & \textbf{Speedup} & \textbf{Acc. Retention} \\
\midrule
\multirow{4}{*}{\textbf{AI2D}}
 & LLaDA-V (32-step) & 500 samples & 73.00 & 49.194 & 1.00$\times$ & 100.0\% \\
 & \quad + RedVTP ($r{=}0.25$) & 500 samples & 71.60 & 38.319 & 1.28$\times$ & 98.1\% \\
 & LLaDA-V (32-step) & full set (3,088) & 72.73 & 41.94 & 1.00$\times$ & 100.0\% \\
 & \quad + \textbf{DRAMA (ours)} & full set (3,088) & \textbf{73.02} & \textbf{2.57} & \textbf{16.32$\times$} & \textbf{100.4\%} \\
\midrule
\multirow{4}{*}{\textbf{MM-Star}}
 & LLaDA-V (32-step) & 500 samples & 50.00 & 34.663 & 1.00$\times$ & 100.0\% \\
 & \quad + RedVTP ($r{=}0.25$) & 500 samples & 42.40 & 28.718 & 1.21$\times$ & 84.8\% \\
 & LLaDA-V (32-step) & full set (1,500) & 50.27 & 29.29 & 1.00$\times$ & 100.0\% \\
 & \quad + \textbf{DRAMA (ours)} & full set (1,500) & 47.60 & \textbf{3.52} & \textbf{8.32$\times$} & \textbf{94.7\%} \\
\midrule
\multirow{2}{*}{\textbf{Average}}
 & \quad + RedVTP ($r{=}0.25$) & 500 samples & --- & --- & 1.25$\times$ & 91.5\% \\
 & \quad + \textbf{DRAMA (ours)} & full set & --- & --- & \textbf{12.32$\times$} & \textbf{97.6\%} \\
\bottomrule
\end{tabular}%
}
\end{table}

\begin{table}[t]
\centering
\caption{Same-backbone wall-clock speedup comparison with RedVTP on LaViDa. RedVTP reports only $r=75\%$ on LaViDa, since LaViDa's built-in $2\times2$ visual-token pooling already retains 25\% of the original tokens. RedVTP LaViDa numbers are taken from \citet{xu2025redvtp}}; DRAMA runs under 128-step think mode. The AI2D and MME columns are measured on every sample (3,088 and 2,374); the MMBench column is measured on a 100-sample subset against a baseline on the same subset. \colorbox{abrank1}{Best}, \colorbox{abrank2}{2nd best}.
\label{tab:redvtp_lavida}
\begin{tabular}{l c c c}
\toprule
\textbf{Method} & \textbf{AI2D} & \textbf{MME} & \textbf{MMBench} \\
\midrule
\multicolumn{4}{l}{\textit{RedVTP regime (LaViDa)}} \\
\quad LaViDa (baseline) & 1.00$\times$ & 1.00$\times$ & 1.00$\times$ \\
\quad + RedVTP ($r=75\%$) & 1.08$\times$ & 1.06$\times$ & 1.06$\times$ \\
\midrule
\multicolumn{4}{l}{\textit{DRAMA regime (LaViDa, 128-step think mode)}} \\
\quad LaViDa (baseline) & 1.00$\times$ & 1.00$\times$ & 1.00$\times$ \\
\quad DRAMA CK Uni  & 1.80$\times$ & 1.72$\times$ & 1.56$\times$ \\
\quad DRAMA CK Prog & 1.82$\times$ & \cellcolor{abrank2}1.82$\times$ & 1.58$\times$ \\
\quad DRAMA FG Prog & \cellcolor{abrank2}1.88$\times$ & \cellcolor{abrank1}\textbf{1.84$\times$} & 1.69$\times$ \\
\quad DRAMA FG Uni  & \cellcolor{abrank1}\textbf{1.91$\times$} & 1.82$\times$ & 1.68$\times$ \\
\bottomrule
\end{tabular}
\end{table}

\subsubsection{Comparison with D$^{3}$ToM~\cite{chang2026d3tom}}

D3ToM~\cite{chang2026d3tom} accelerates LaViDa by merging visual tokens under decider-token guidance.
It reports wall-clock time only on LLaVA-Bench in a short-generation setting ($T{=}32$, $|O|{=}64$),
so its published speedups do not transfer to think mode. We therefore re-run D3ToM-t at its own
default operating point ($25\%$ of visual tokens retained, merged at layer~3) under our 128-step
think mode and against our own baseline (Table~\ref{tab:d3tom_lavida}). It gains only
$1.15$--$1.23\times$ against DRAMA's $1.63$--$1.91\times$, and retains less accuracy on every
benchmark, falling to $0.81$ and $0.83$ on MM-Star and SEED-Bench where DRAMA holds
$0.91$--$0.96$. On AI2D and MMMU, the two benchmarks D3ToM also evaluates, our reproduction
reaches $0.96$ and $0.94$ against the $0.94$--$0.96$ and $0.97$--$0.98$ they report.

\begin{table*}[t]
\centering
\caption{Comparison with D3ToM on LaViDa. All rows are re-run by us under 128-step think mode, D3ToM included, so that accuracy and speedup are measured under one regime against one baseline; retention is accuracy divided by that baseline. Rows use the full set for AI2D, MME, MMMU, MMstar, and a 1000-sample subset for SEED-Bench. \colorbox{abrank1}{Best}, \colorbox{abrank2}{2nd best}.}
\label{tab:d3tom_lavida}
\resizebox{\textwidth}{!}{%
\begin{tabular}{l cc cc cc cc c}
\toprule
 & \multicolumn{2}{c}{\textbf{AI2D}} & \multicolumn{2}{c}{\textbf{MMMU}} & \multicolumn{2}{c}{\textbf{MM-Star}} & \multicolumn{2}{c}{\textbf{SEED-Bench}} & \textbf{Avg.} \\
\cmidrule(lr){2-3} \cmidrule(lr){4-5} \cmidrule(lr){6-7} \cmidrule(lr){8-9} \cmidrule(lr){10-10}
\textbf{Method} & Acc (Ret.) & Sp. & Acc (Ret.) & Sp. & Acc (Ret.) & Sp. & Acc (Ret.) & Sp. & Sp. \\
\midrule
LaViDa (baseline) & 69.88 (1.00) & 1.00$\times$ & 42.44 (1.00) & 1.00$\times$ & 43.00 (1.00) & 1.00$\times$ & 63.50 (1.00) & 1.00$\times$ & 1.00$\times$ \\
D3ToM-t ($keep{=}0.25$, L3) & 66.80 (0.96) & 1.20$\times$ & 39.89 (0.94) & 1.15$\times$ & 38.80 (0.81) & 1.21$\times$ & 53.00 (0.83) & 1.23$\times$ & 1.20$\times$ \\
\midrule
DRAMA CK Uni & 67.45 (0.97) & 1.80$\times$ & \cellcolor{abrank1}\textbf{42.00 (0.99)} & 1.70$\times$ & \cellcolor{abrank1}\textbf{40.00 (0.93)} & 1.65$\times$ & \cellcolor{abrank1}\textbf{61.00 (0.96)} & 1.77$\times$ & 1.73$\times$ \\
DRAMA CK Prog & \cellcolor{abrank1}\textbf{67.81 (0.97)} & 1.82$\times$ & \cellcolor{abrank2}41.44 (0.98) & 1.72$\times$ & 39.00 (0.91) & 1.63$\times$ & 59.10 (0.93) & 1.78$\times$ & 1.74$\times$ \\
DRAMA FG Prog & 67.58 (0.97) & \cellcolor{abrank2}1.88$\times$ & 41.11 (0.97) & \cellcolor{abrank1}\textbf{1.80$\times$} & \cellcolor{abrank2}39.13 (0.91) & \cellcolor{abrank1}\textbf{1.76$\times$} & 59.10 (0.93) & \cellcolor{abrank1}\textbf{1.91$\times$} & \cellcolor{abrank1}\textbf{1.84$\times$} \\
DRAMA FG Uni & \cellcolor{abrank1}\textbf{67.81 (0.97)} & \cellcolor{abrank1}\textbf{1.91$\times$} & 41.22 (0.97) & \cellcolor{abrank2}1.75$\times$ & \cellcolor{abrank1}\textbf{40.00 (0.93)} & \cellcolor{abrank2}1.72$\times$ & \cellcolor{abrank1}\textbf{61.00 (0.96)} & \cellcolor{abrank2}1.89$\times$ & \cellcolor{abrank2}1.82$\times$ \\
\bottomrule
\end{tabular}%
}
\end{table*}

\subsection{Generalization to Additional Benchmarks}
\label{sec:appendix_extra}
\begin{table}[t]
\centering
\caption{Held-out evaluation on POPE and TextVQA, both on LLaDA-V. Neither benchmark was used when setting any threshold. Speedup is wall-clock against the 64-step baseline of the same benchmark. POPE reports F1 on 500 samples; TextVQA reports accuracy on 100 samples. ``SGAS only'' is DRAMA with SVTP disabled.}
\label{tab:extra}
\begin{tabular}{l cc cc}
\toprule
 & \multicolumn{2}{c}{\textbf{POPE} (F1)} & \multicolumn{2}{c}{\textbf{TextVQA} (Acc)} \\
\cmidrule(lr){2-3}\cmidrule(lr){4-5}
\textbf{Setting} & Score & Speedup & Score & Speedup \\
\midrule
Baseline (64 steps) & 88.30 & 1.00$\times$ & 64.60 & 1.00$\times$ \\
Baseline (32 steps) & 86.37 & 1.90$\times$ & 60.70 & 2.00$\times$ \\
Baseline (16 steps) & 87.24 & 3.77$\times$ & -- & -- \\
\midrule
SGAS only (FG Uni)  & 88.30 & 3.81$\times$ & -- & -- \\
DRAMA FG Prog       & 85.35 & 10.69$\times$ & 51.60 & 7.91$\times$ \\
DRAMA FG Uni        & 85.35 & 11.23$\times$ & 52.50 & 5.87$\times$ \\
\bottomrule
\end{tabular}
\end{table}
Neither TextVQA~\cite{singh2019towards} nor POPE~\cite{li2023evaluating} was used when setting any threshold, so both serve as held-out checks on whether the redundancy structures of Section~\ref{sec:spatial}~\ref{sec:temporal}
transfer beyond the benchmarks we report in the main paper. They probe opposite ends of the same question. POPE asks whether a named object is present, which depends on the visual prefix surviving pruning; TextVQA asks open-ended questions about dense text inside the image, which depends on both the prefix and a sufficient number of denoising steps. Table~\ref{tab:extra} reports both.

On POPE, SGAS alone is lossless: all four schedules reproduce the 64-step F1 of 88.30 while reaching up to $3.81\times$, confirming that the temporal stabilization measured in Section~\ref{sec:temporal} holds on a task the thresholds were never tuned for. Adding SVTP costs $2.95$ F1 and buys $11.23\times$ speedup. That the loss appears only once the visual prefix is pruned is consistent with the nature of the task: object-existence judgements read directly off the prefix, so they are the first to suffer when it is compressed.

TextVQA is where DRAMA is weakest: accuracy falls from $64.60$ at 64 steps to $52.50$ for FG Uni at $5.87\times$, and halving the step budget alone already accounts for $3.90$ of those points. As stated in Appendix~\ref{sec:appendix_limitations}, DRAMA adds no mechanism to recover quality lost to a reduced budget, and on open-ended text-dense inputs that cost is simply larger than on the six main benchmarks.
\section{Derivation of the Output-Stationarity Bound}
\label{app:bound}

This appendix supplies the derivation of the Pinsker-type output-stationarity bound invoked in \cref{sec:stopping}, showing that the per-step layer-output drift vanishes as a direct corollary of the JSD collapse identified in \cref{sec:temporal}.

\subsection{Setup and Decomposition}
At denoising step $t$ and transformer layer $l$, let $\mathbf{A}^{(t)}_l \!\in\! \mathbb{R}^{H \times N \times N}$ be the attention tensor, $\mathbf{V}^{(t)}_l \!\in\! \mathbb{R}^{H \times N \times d_v}$ the value tensor, and $\mathbf{O}^{(t)}_l = \mathbf{A}^{(t)}_l \mathbf{V}^{(t)}_l$ the layer output. Adding and subtracting $\mathbf{A}^{(t)}_l \mathbf{V}^{(t-1)}_l$ yields the canonical decomposition
\begin{equation}
    \Delta\mathbf{O}^{(t)}_l
    \;\equiv\; \mathbf{O}^{(t)}_l \!-\! \mathbf{O}^{(t-1)}_l
    \;=\;
    \underbrace{\Delta\mathbf{A}^{(t)}_l \mathbf{V}^{(t-1)}_l}_{\text{attention drift}}
    \;+\;
    \underbrace{\mathbf{A}^{(t)}_l \Delta\mathbf{V}^{(t)}_l}_{\text{value update}},
    \label{eq:app_decomp}
\end{equation}
with $\Delta\mathbf{A}^{(t)}_l \!=\! \mathbf{A}^{(t)}_l \!-\! \mathbf{A}^{(t-1)}_l$ and $\Delta\mathbf{V}^{(t)}_l \!=\! \mathbf{V}^{(t)}_l \!-\! \mathbf{V}^{(t-1)}_l$. The triangle inequality, combined with submultiplicativity in the Frobenius/operator-norm pair, gives
\begin{equation}
    \bigl\| \Delta\mathbf{O}^{(t)}_l \bigr\|_F
    \;\leq\;
    \bigl\| \Delta\mathbf{A}^{(t)}_l \bigr\|_F \, \bigl\| \mathbf{V}^{(t-1)}_l \bigr\|_{\mathrm{op}}
    \;+\;
    \bigl\| \mathbf{A}^{(t)}_l \bigr\|_{\mathrm{op}} \, \bigl\| \Delta\mathbf{V}^{(t)}_l \bigr\|_F.
    \label{eq:app_triangle}
\end{equation}
The two summands are bounded separately in the next two subsections.

\subsection{Bounding the Attention-Drift Term via JSD}
For each (head $h$, query $i$), the per-row attention vectors $\mathbf{a}^{(t)}_{l,h,i}, \mathbf{a}^{(t-1)}_{l,h,i} \!\in\! \Delta^{N-1}$ are probability distributions over key positions. They satisfy the Lin (1991) inequality, the JSD analogue of Pinsker's bound,
\begin{equation}
    \mathrm{JSD}\bigl(\mathbf{a}^{(t)}_{l,h,i} \,\|\, \mathbf{a}^{(t-1)}_{l,h,i}\bigr)
    \;\geq\; \tfrac{1}{2}\, \mathrm{TV}\bigl(\mathbf{a}^{(t)}_{l,h,i}, \mathbf{a}^{(t-1)}_{l,h,i}\bigr)^{2}.
    \label{eq:app_lin}
\end{equation}
Combined with $\| \mathbf{a}^{(t)}_{l,h,i} \!-\! \mathbf{a}^{(t-1)}_{l,h,i} \|_2 \leq \| \cdot \|_1 = 2\,\mathrm{TV}(\cdot)$ and aggregation over all rows,
\begin{equation}
    \bigl\| \Delta\mathbf{A}^{(t)}_l \bigr\|_F^{2}
    \;=\; \sum_{h,i} \bigl\| \mathbf{a}^{(t)}_{l,h,i} \!-\! \mathbf{a}^{(t-1)}_{l,h,i} \bigr\|_2^{2}
    \;\leq\; 8\,HN \, \bar{J}_l(t),
    \label{eq:app_dA}
\end{equation}
where $\bar{J}_l(t)$ is the mean per-layer divergence defined in Eq.~\eqref{eq:mean_jsd}. Substituting Eq.~\eqref{eq:app_dA} into the first summand of Eq.~\eqref{eq:app_triangle} recovers the headline form used in the main text,
\begin{equation}
    \bigl\| \Delta\mathbf{A}^{(t)}_l \mathbf{V}^{(t-1)}_l \bigr\|_F
    \;\leq\; C_l \, \sqrt{2\, \bar{J}_l(t)},
    \qquad
    C_l \;:=\; 2\sqrt{HN} \cdot \bigl\| \mathbf{V}^{(t-1)}_l \bigr\|_{\mathrm{op}}.
    \label{eq:app_drift}
\end{equation}
Standard layer-norm and weight-decay regularisation keep $\| \mathbf{V}^{(t-1)}_l \|_{\mathrm{op}}$ in a bounded operator-norm regime across steps, so $C_l$ is a step-independent constant for the purposes of the argument.

\subsection{Bounding the Value-Update Term via Layer-wise Propagation}
The value tensor at layer $l$ is computed from the previous layer's output through a Lipschitz block $f_l$ (linear projection composed with layer norm and, in pre-norm transformers, residual addition): $\mathbf{V}^{(t)}_l = f_l(\mathbf{O}^{(t)}_{l-1})$ with Lipschitz constant $L_l < \infty$. Hence
\begin{equation}
    \bigl\| \Delta\mathbf{V}^{(t)}_l \bigr\|_F
    \;\leq\; L_l \, \bigl\| \Delta\mathbf{O}^{(t)}_{l-1} \bigr\|_F.
    \label{eq:app_vlip}
\end{equation}
Inserting Eqs.~\eqref{eq:app_drift} and \eqref{eq:app_vlip} into Eq.~\eqref{eq:app_triangle} produces the layer-wise recurrence
\begin{equation}
    \bigl\| \Delta\mathbf{O}^{(t)}_l \bigr\|_F
    \;\leq\;
    C_l \sqrt{2\,\bar{J}_l(t)}
    \;+\;
    \rho_l \, \bigl\| \Delta\mathbf{O}^{(t)}_{l-1} \bigr\|_F,
    \qquad
    \rho_l \;:=\; \bigl\| \mathbf{A}^{(t)}_l \bigr\|_{\mathrm{op}} L_l.
    \label{eq:app_recur}
\end{equation}
Unrolling along the layer stack with the boundary condition $\Delta\mathbf{O}^{(t)}_0 \!=\! \mathbf{0}$ at the input embedding (whose only step-to-step change comes from the ${\sim}N_g/T$ newly unmasked tokens, which the converged attention assigns negligible aggregate weight) gives
\begin{equation}
    \bigl\| \Delta\mathbf{O}^{(t)}_l \bigr\|_F
    \;\leq\;
    \sum_{k=1}^{l} \Bigl( \prod_{j=k+1}^{l} \rho_j \Bigr) \, C_k \sqrt{2\,\bar{J}_k(t)}.
    \label{eq:app_unroll}
\end{equation}
The temporal dissection in \cref{sec:temporal} establishes that $\bar{J}_k(t) \!\to\! 0$ \emph{simultaneously across all $k \!\in\! [L]$} once past the structure-formation phase (the bright plateau region of \cref{fig:temporal}(b)). Every term in the sum is therefore suppressed in lockstep, and $\| \Delta\mathbf{O}^{(t)}_l \|_F \!\to\! 0$ uniformly in $l$. The simultaneity is essential: it converts what would otherwise be a worst-case accumulating bound into one in which all contributing factors vanish concurrently.

\subsection{From Layer Output to Decoded Sequence}
The unembedding head is a fixed linear map $\mathbf{W}_U$ followed by softmax, so the per-token logit drift obeys $\| \Delta \mathbf{z}^{(t)} \|_F \leq \| \mathbf{W}_U \|_{\mathrm{op}} \, \| \Delta\mathbf{O}^{(t)}_L \|_F \!\to\! 0$. Combined with the dLLM low-confidence remasking schedule, which unmasks only ${\sim}N_g/T$ positions per step, the argmax-decoded sequence inherits the same stationarity and tends to stabilize before the nominal budget $T$ is exhausted. This justifies the promotion of the decoded text to a sufficiency surrogate in \cref{sec:stopping}, replacing any runtime monitoring of $\bar{J}_l(t)$ that would otherwise require the per-layer attention caching this work seeks to eliminate.

\section{Additional Redundancy Analysis and Results}
\label{sec:appendix_redun}
\subsection{Dissecting Spatial Redundancy: Analysis per Benchmark}
\label{sec:appendix_spatial}
We visualize spatial redundancy with MMMU~\cite{yue2024mmmu} for illustration in the main paper. Below, we provide the same four-panel spatial analysis for all six benchmarks individually. Each figure reports: (a)~per-rank crop attention fraction, (b)~a representative spatial attention heatmap, (c)~token-level concentration CDF stratified by modality pair, and (d)~the per-query visual token budget $\kappa(0.9)$ across denoising steps.

The key findings from the main paper generalize consistently: attention mass concentrates on 1--2 dominant patches across all benchmarks, the t2v concentration curve is the steepest among all modality pairs, and $\kappa(0.9)$ remains step-invariant. Benchmarks with simpler visual content (e.g., AI2D~\cite{kembhavi2016diagram} diagrams, MME~\cite{fu2023mme} perceptual images) exhibit slightly higher concentration, while those with richer scenes (e.g., SEEDBench~\cite{li2023seed}, MMBench~\cite{liu2024mmbench}) show moderately broader spatial distributions---yet even in the latter case, the most-attended patch captures ${\geq}40\%$ of output$\to$image attention.
\begin{figure*}[t]
    \centering
    \includegraphics[width=\textwidth]{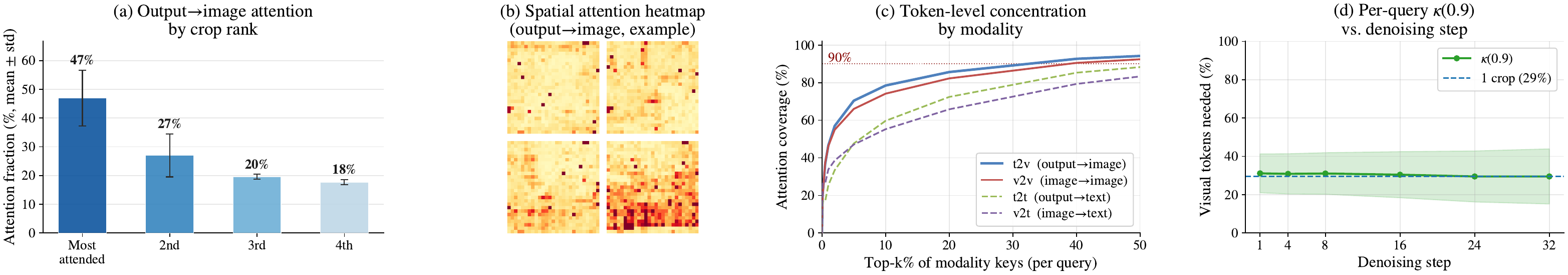}
    \caption{\textbf{Spatial redundancy analysis -- AI2D.} The diagram-centric nature of AI2D leads to highly concentrated attention on a single informative patch.}
    \label{fig:spatial_ai2d}
\end{figure*}

\begin{figure*}[t]
    \centering
    \includegraphics[width=\textwidth]{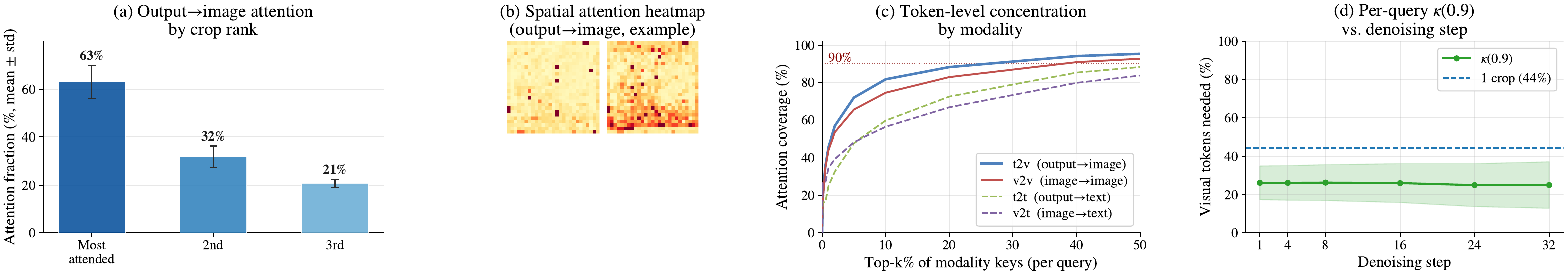}
    \caption{\textbf{Spatial redundancy analysis -- MMBench.} Despite diverse image content, the top 2--3 patches consistently capture the majority of output$\to$image attention.}
    \label{fig:spatial_mmbench}
\end{figure*}

\begin{figure*}[t]
    \centering
    \includegraphics[width=\textwidth]{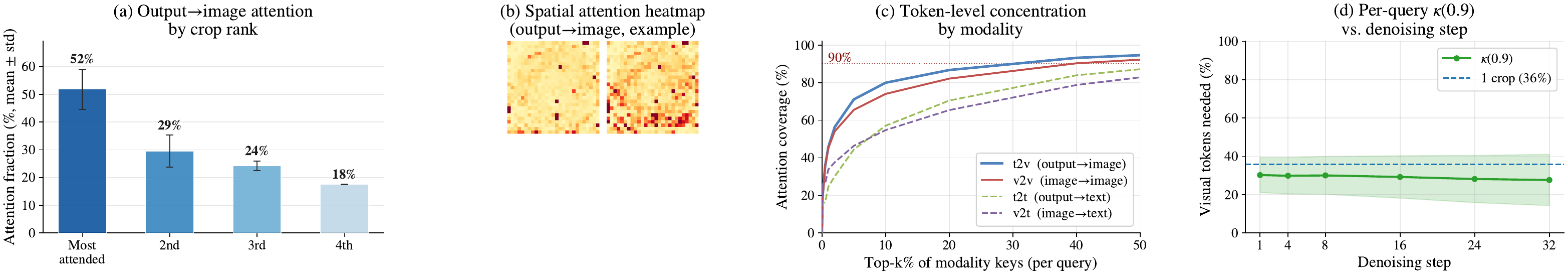}
    \caption{\textbf{Spatial redundancy analysis -- MME.} Perceptual tasks in MME yield concentrated attention patterns, with $\kappa(0.9)$ remaining stable across all denoising steps.}
    \label{fig:spatial_mme}
\end{figure*}

\begin{figure*}[t]
    \centering
    \includegraphics[width=\textwidth]{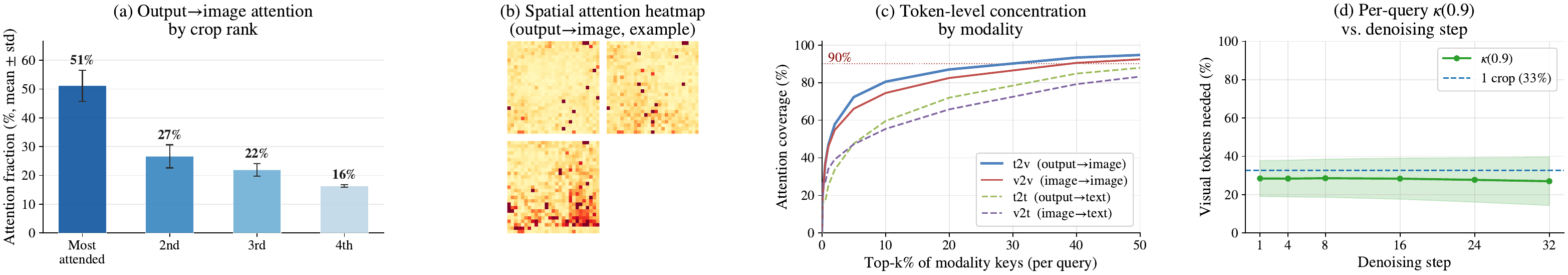}
    \caption{\textbf{Spatial redundancy analysis -- MM-Star.} Knowledge-intensive queries in MM-Star show similar spatial concentration to MMMU.}
    \label{fig:spatial_mmstar}
\end{figure*}

\begin{figure*}[t]
    \centering
    \includegraphics[width=\textwidth]{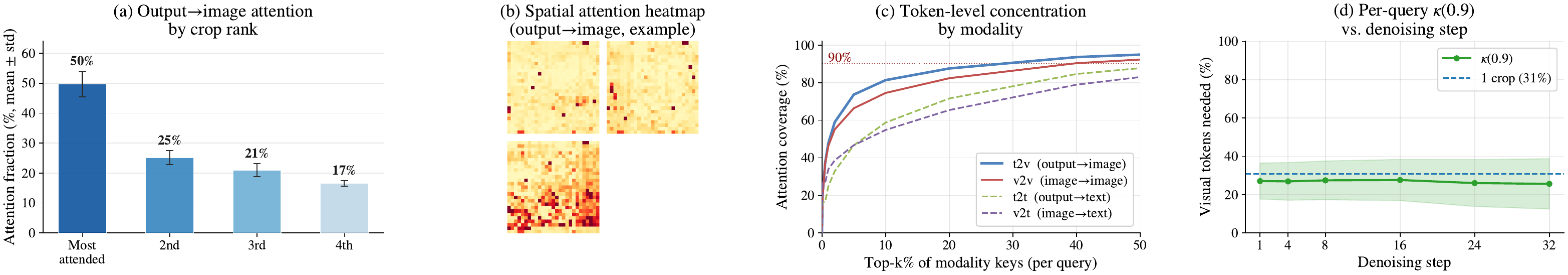}
    \caption{\textbf{Spatial redundancy analysis -- SEED-Bench.} SEED-Bench exhibits slightly broader spatial distributions due to its diverse scene understanding tasks, yet the dominant patch still captures ${\geq}40\%$ of attention.}
    \label{fig:spatial_seedbench}
\end{figure*}
\subsection{Dissecting Temporal Redundancy: Analysis per Benchmark}
\label{sec:appendix_temporal}
We present the temporal analysis averaged over all benchmarks in the main paper. Here we provide per-benchmark breakdowns, each showing (a)~the per-layer inter-step JSD $\bar{J}_l(t)$ across all 64 denoising steps, and (b)~the layer$\times$step redundancy heatmap $R_l(t)$.

The three-phase temporal structure identified in the main paper---structure formation, stable plateau, and terminal refinement---is clearly visible in every benchmark. Across all datasets, the JSD peaks within the first 2--5 steps, then drops by over two orders of magnitude into a prolonged plateau where $R_l(t) > 0.99$, before a secondary rise in the final 5--10 steps. The heatmaps further reveal that deeper layers (L25--L32) exhibit the lowest redundancy scores throughout the trajectory, consistent with these layers performing the fine-grained disambiguation required for final token selection.

Notably, the plateau onset and duration are remarkably consistent across benchmarks despite their diverse task characteristics (diagram understanding in AI2D~\cite{kembhavi2016diagram}, perceptual reasoning in MME~\cite{fu2023mme}, multi-discipline knowledge in MMMU~\cite{yue2024mmmu}), confirming that the temporal redundancy is a structural property of the denoising dynamics rather than a task-specific artefact. This consistency is what enables a single set of SGAS hyperparameters to work effectively across all benchmarks without per-task tuning.

\begin{figure*}[t]
    \centering
    \includegraphics[width=\textwidth]{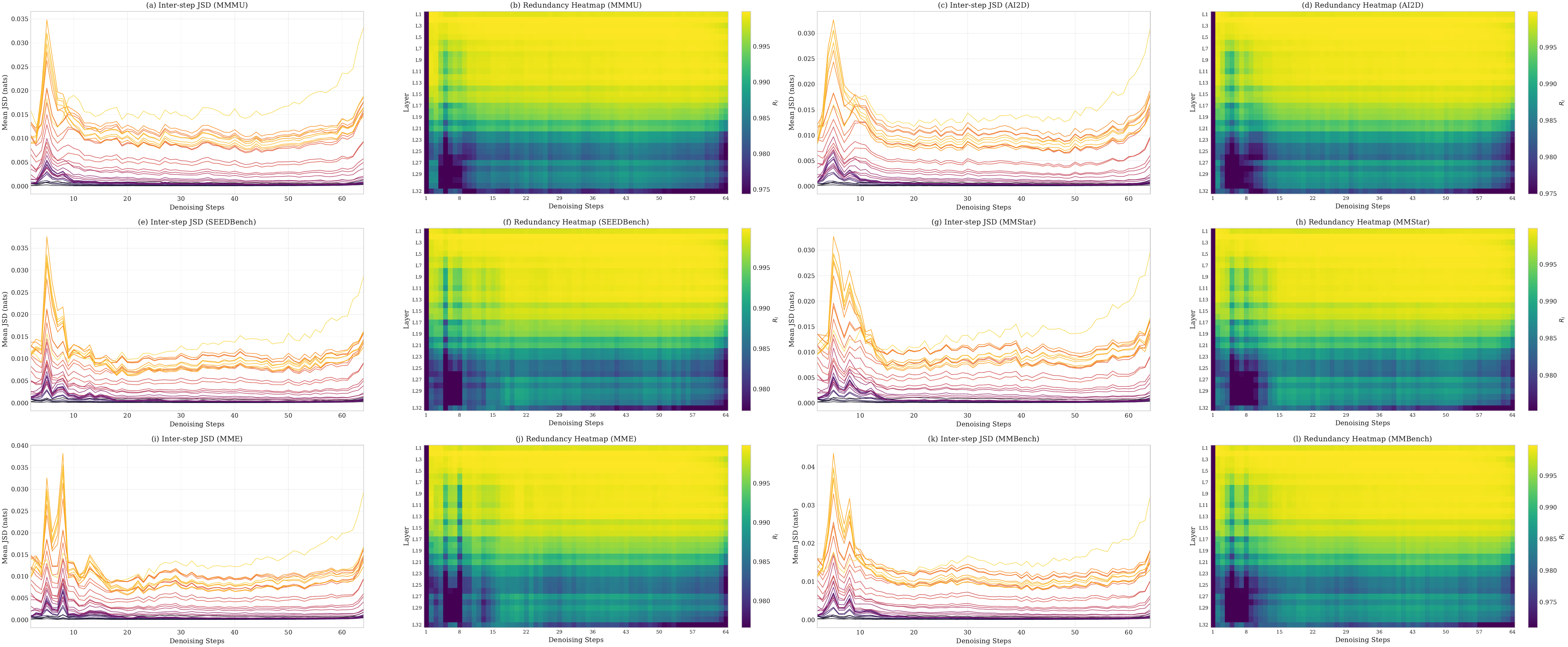}
    \caption{\textbf{Temporal redundancy analysis for all benchmarks.} (a)~Per-layer inter-step JSD across denoising steps. (b)~Layer$\times$step redundancy heatmap.}
    \label{fig:temporal_all}
\end{figure*}
\begin{figure}[t]
    \centering
    \includegraphics[width=\linewidth]{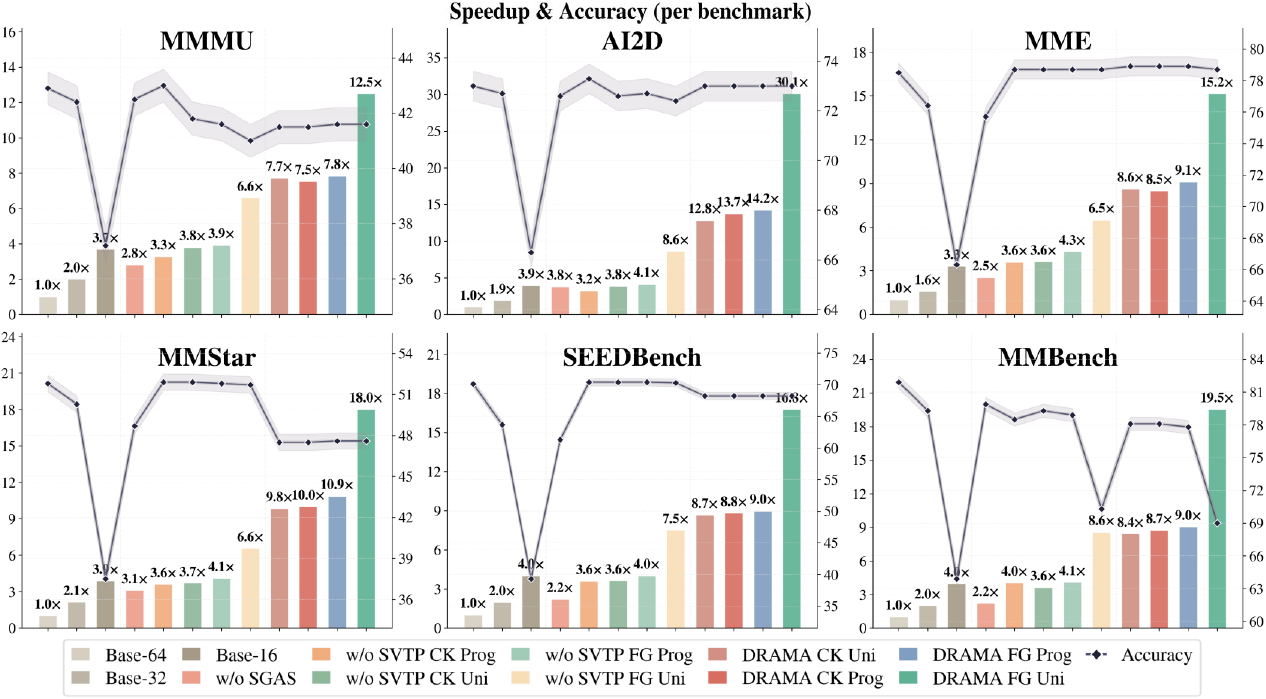}
    \caption{DRAMA achieves an average $18.67\times$ inference speedup over fixed-budget baselines across six benchmarks (MMMU~\cite{yue2024mmmu}, AI2D~\cite{kembhavi2016diagram}, MME~\cite{fu2023mme}, MM-Star~\cite{chen2024we}, SEED-Bench~\cite{li2023seed}, and MMBench~\cite{liu2024mmbench}) while maintaining competitive accuracy.}
    \label{fig:teaser}
\end{figure}
\subsection{Pareto Frontier}
\label{sec:appendix_pareto}
\begin{figure}[h]
    \centering
    \includegraphics[width=\linewidth]{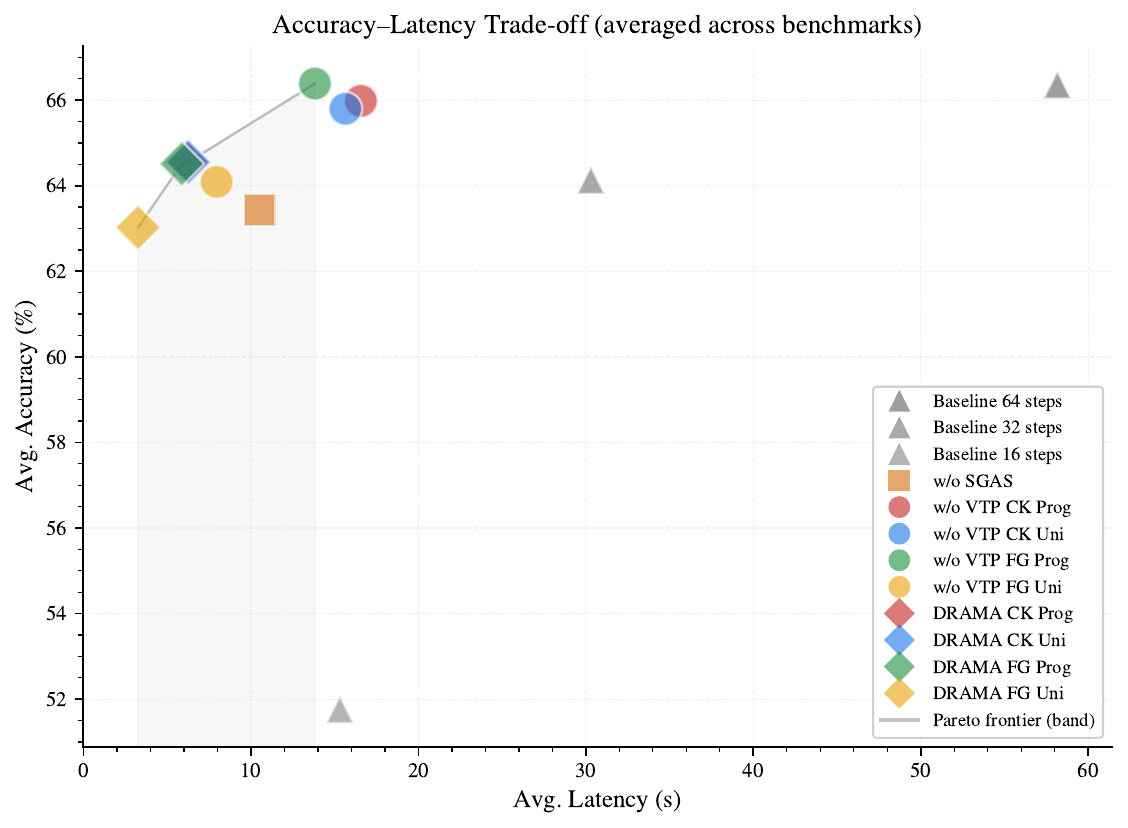}
    \caption{\textbf{Averaged accuracy--latency Pareto frontier} across all benchmarks. Each point represents a method's mean accuracy and mean latency. DRAMA variants occupy the Pareto front, achieving superior accuracy at substantially lower latency than fixed-budget baselines.}
    \label{fig:pareto_avg}
\end{figure}

We present the accuracy--latency Pareto frontier averaged across all six benchmarks in \Cref{fig:pareto_avg}. The averaged view confirms that DRAMA variants consistently dominate the frontier: no fixed-budget baseline achieves a comparable combination of speed and accuracy. Among DRAMA variants, CK~Prog and CK~Uni occupy the high-accuracy region (${\sim}64.5\%$, ${\sim}6$\,s), while FG~Uni pushes to the extreme-speed end (${\sim}63.0\%$, ${\sim}3.3$\,s), demonstrating the flexibility of the framework.
\subsection{Step Distribution Analysis per Benchmark}
\label{sec:appendix_step}
We provide the combined step distribution overview in the main paper. Below we show per-benchmark distributions with full statistics for each of the four SGAS configurations under both the w/o~SVTP ablation (top row) and the full DRAMA setting (bottom row).

Each subplot displays a histogram of actual stopping steps overlaid with a kernel density estimate (KDE), together with summary statistics: mean ($\mu$), median, standard deviation ($\sigma$), computational saving relative to the 64-step baseline, and sample count ($N$). Several patterns are consistent across benchmarks:
\begin{itemize}
    \item \textbf{DRAMA} (bottom rows) shows similar stopping patterns to the w/o~SVTP ablation (top rows), confirming that SVTP and SGAS operate on orthogonal axes (spatial vs.\ temporal) and compose independently.
    \item \textbf{w/o SVTP FG~Uni} concentrates exits at the earliest steps (typically around 7$--$11 steps, with a later mean of 15.6 on MM-Star), reflecting its willingness to terminate as soon as both criteria are met, resulting in the highest speedup but occasionally at the cost of accuracy on harder benchmarks.
    \item \textbf{w/o SVTP CK variants} cluster around the first eligible checkpoint (step~16), as the sparse evaluation schedule prevents earlier termination regardless of output convergence.
    \item \textbf{w/o SVTP Prog Variants} consistently shifts the distribution rightward compared to uniform scheduling, reflecting its stricter early-step requirements that act as a safeguard during the structure-formation phase.
\end{itemize}

The alignment between the stopping step distributions and the JSD plateau onset provides empirical validation that SGAS terminates denoising precisely when the temporal redundancy analysis predicts the output has stabilized.
\begin{figure*}[t]
    \centering
    \includegraphics[width=\textwidth]{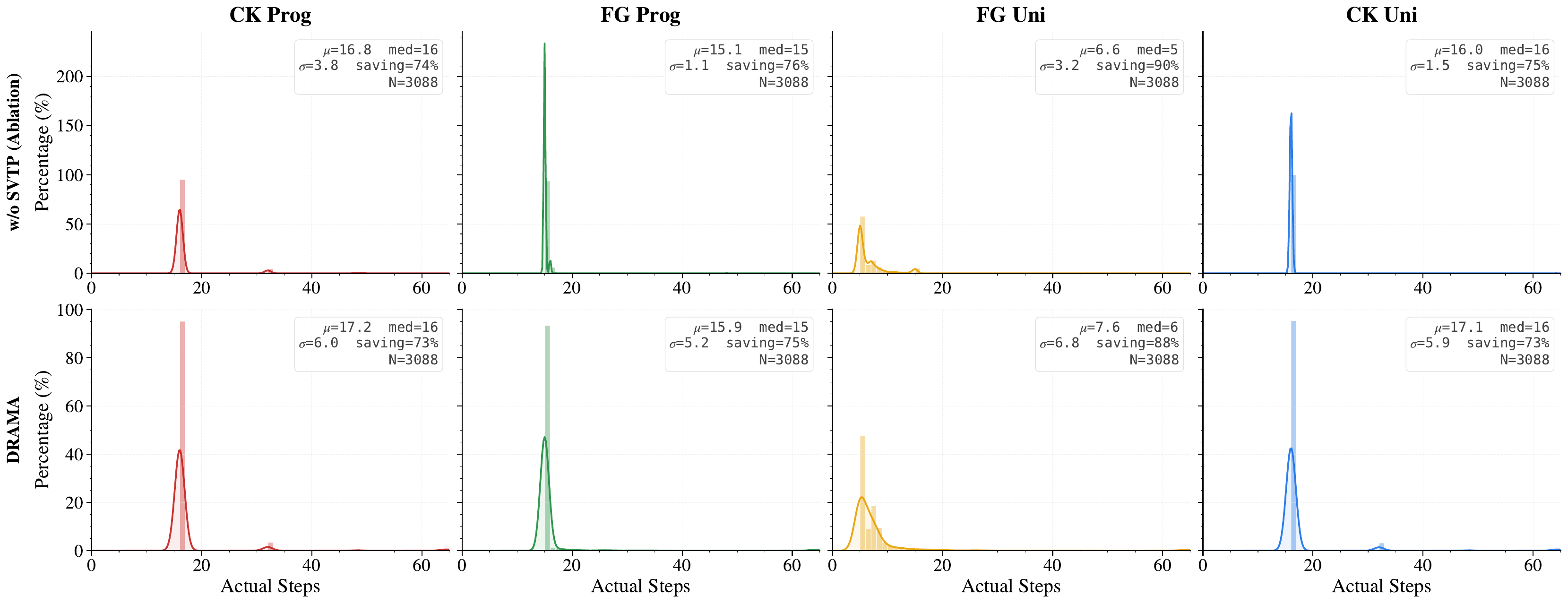}
    \caption{\textbf{Step distribution -- AI2D.} FG~Uni achieves $85\%$/$79\%$ step saving (w/o SVTP / DRAMA) owing to the concentrated spatial attention on diagram elements.}
    \label{fig:step_ai2d}
\end{figure*}

\begin{figure*}[t]
    \centering
    \includegraphics[width=\textwidth]{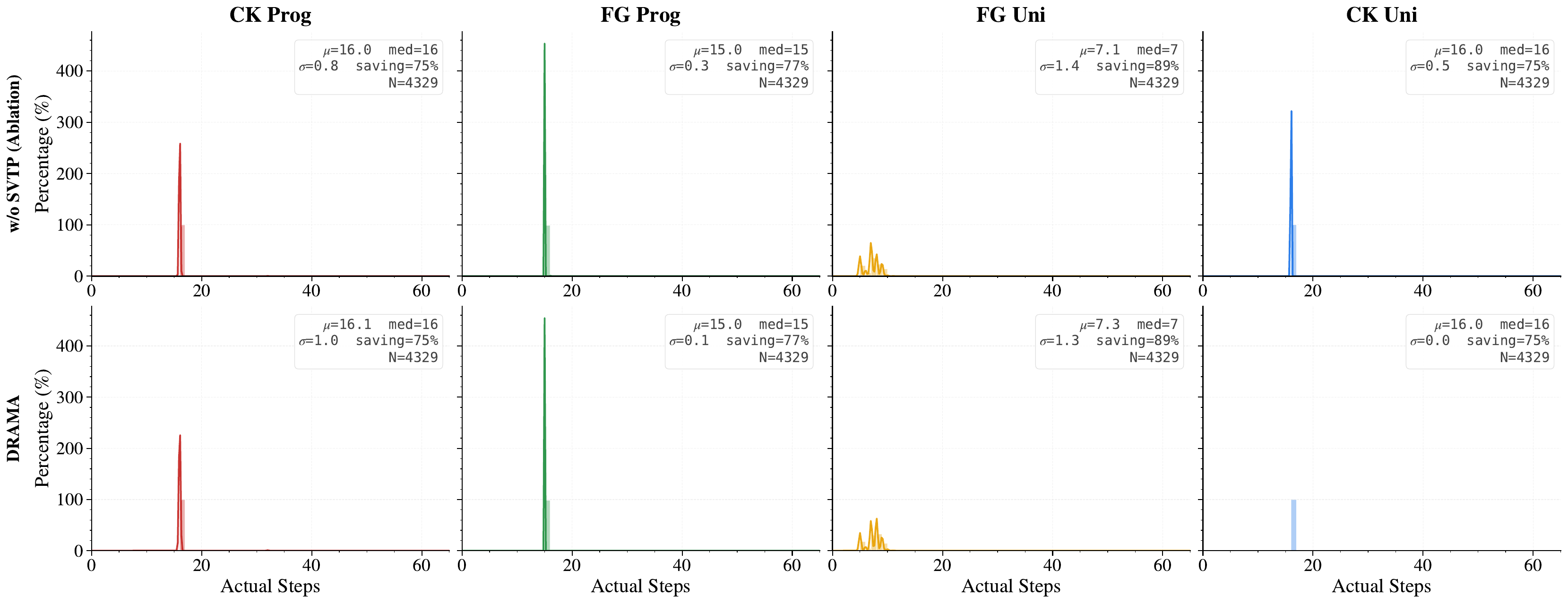}
    \caption{\textbf{Step distribution -- MMBench.} The diverse visual content of MMBench leads to a slightly broader stopping distribution compared to simpler benchmarks.}
    \label{fig:step_mmbench}
\end{figure*}

\begin{figure*}[t]
    \centering
    \includegraphics[width=\textwidth]{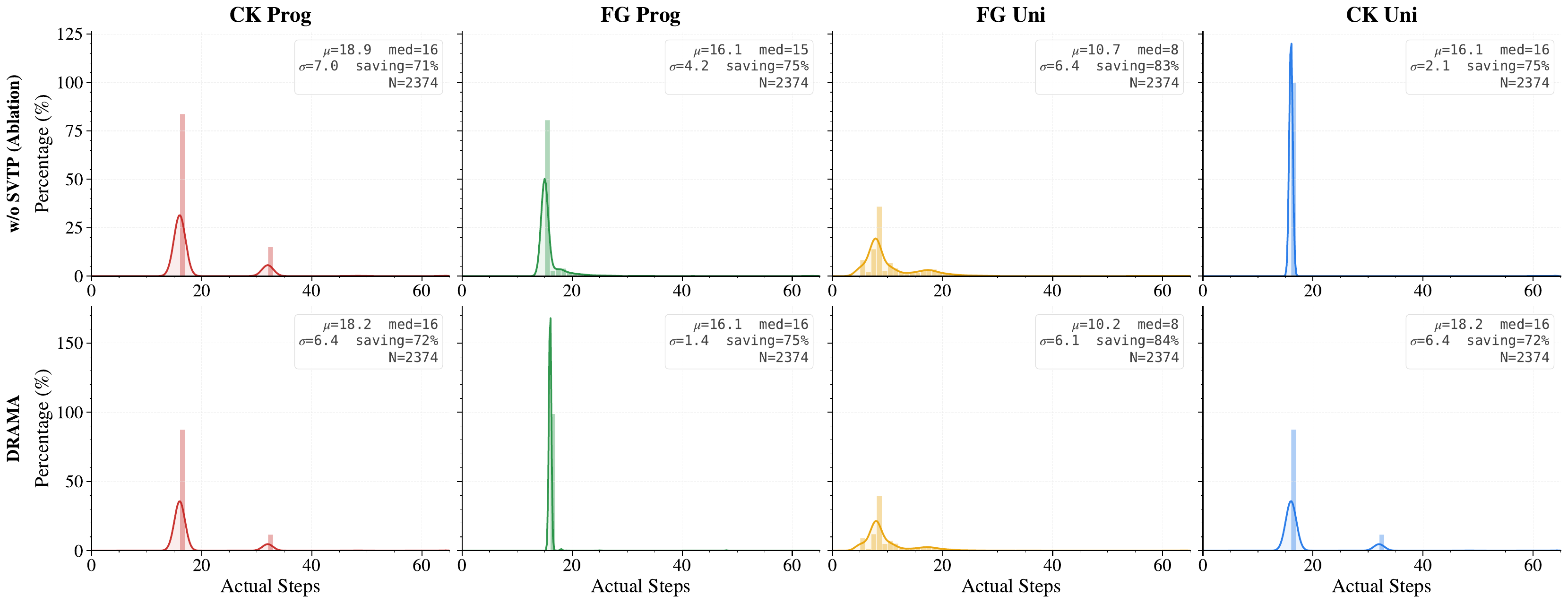}
    \caption{\textbf{Step distribution -- MME.} Perceptual yes/no tasks converge rapidly, with most samples stopping within the first 15--20 steps.}
    \label{fig:step_mme}
\end{figure*}

\begin{figure*}[t]
    \centering
    \includegraphics[width=\textwidth]{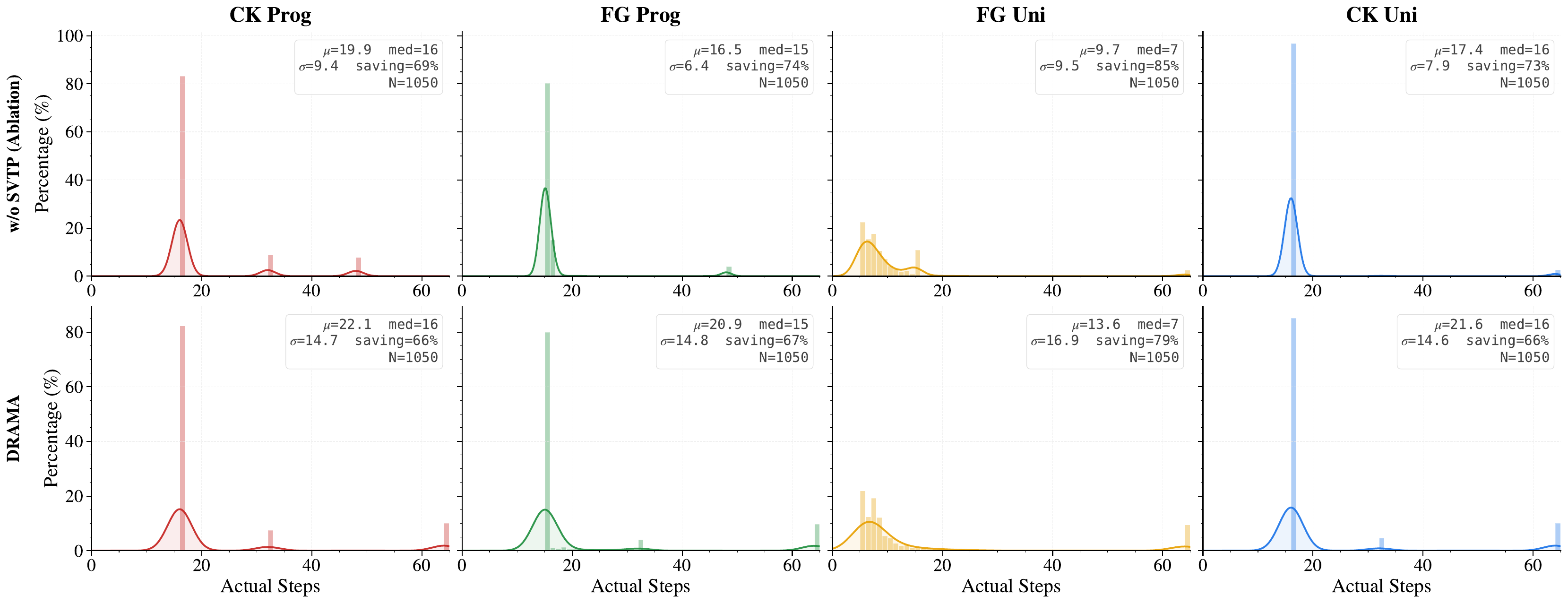}
    \caption{\textbf{Step distribution -- MMMU.} Multi-discipline knowledge questions exhibit moderate step savings, with progressive scheduling providing a useful accuracy safeguard.}
    \label{fig:step_mmmu}
\end{figure*}

\begin{figure*}[t]
    \centering
    \includegraphics[width=\textwidth]{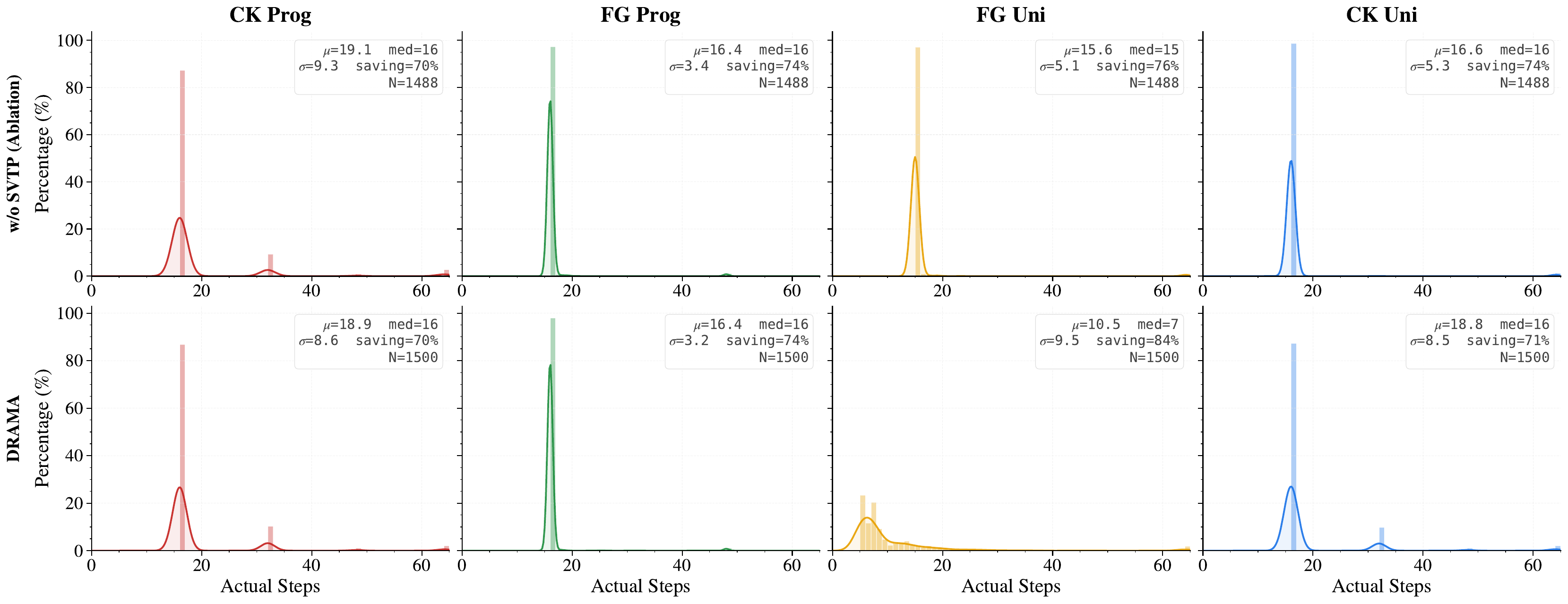}
    \caption{\textbf{Step distribution -- MM-Star.} Knowledge-intensive queries show stopping patterns similar to MMMU, confirming that task difficulty correlates with stopping behaviour.}
    \label{fig:step_mmstar}
\end{figure*}

\begin{figure*}[t]
    \centering
    \includegraphics[width=\textwidth]{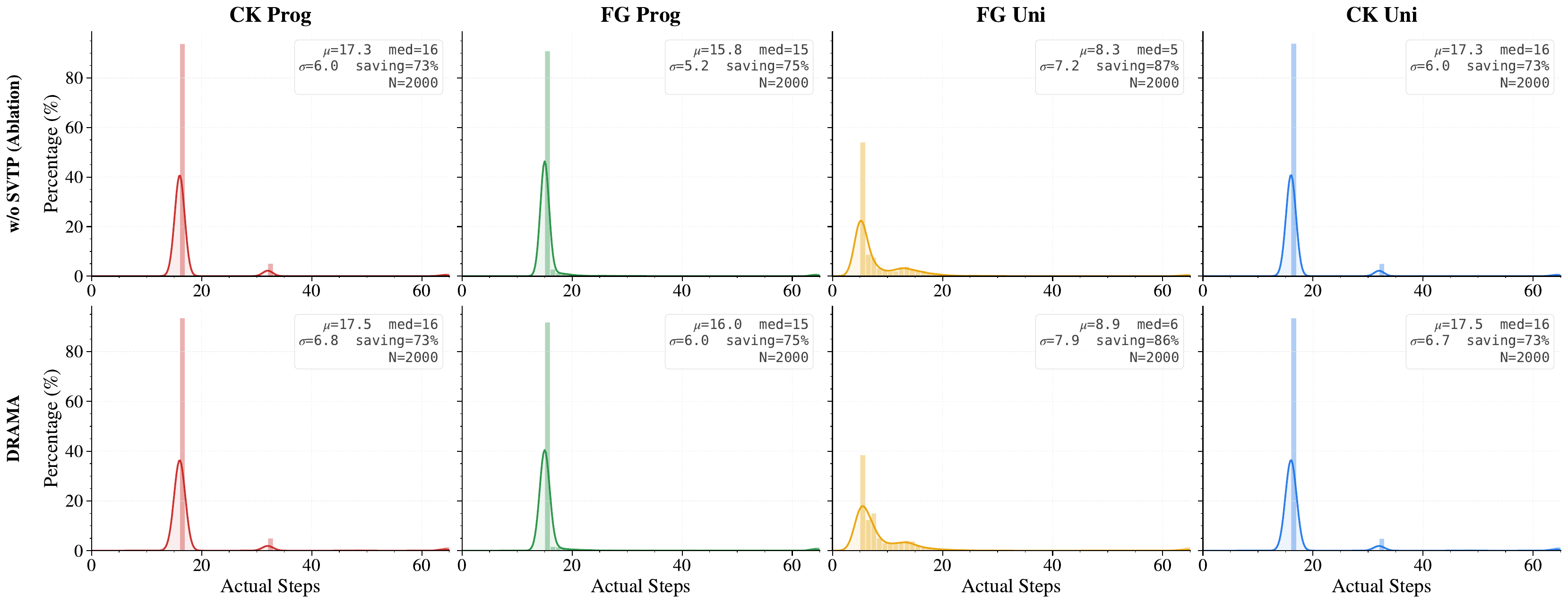}
    \caption{\textbf{Step distribution -- SEED-Bench.} The diverse scene understanding tasks in SEED-Bench produce a broad stopping distribution, yet SGAS still achieves substantial step savings.}
    \label{fig:step_seedbench}
\end{figure*}

\section{Additional Related Work}
\subsection{Attention-Based Acceleration in Vision and Language Models}
Attention-based acceleration has been explored in autoregressive vision-language models (VLMs) and generative diffusion transformers (DiTs), typically following an analyze-then-accelerate paradigm: first identifying structured attention patterns, then designing targeted strategies to exploit them.
In autoregressive VLMs, methods such as FlexPrefill~\cite{lai2025flexprefill} and MInference~\cite{jiang2024minference, li2025mminference} observe that attention is sparse and concentrated on a few key tokens, motivating attention sparsification and token pruning.
In generative DiTs, DiTFastAttn~\cite{yuan2024ditfastattn, zhang2025ditfastattnv2} identifies temporal similarity in attention maps across denoising steps and caches redundant computations accordingly.
Despite the success of this paradigm in both domains, no systematic redundancy analysis-grounded acceleration  or attention-based acceleration framework exists for multimodal dLLMs. Our work addresses this gap by dissecting attention redundancy across two orthogonal axes, spatial and temporal, and deriving dimension-specific acceleration strategies from the empirical findings.
\section{Preliminary}
\label{sec:preliminary}
\subsection{Diffusion Large Language Models.}
Diffusion large language models (dLLMs)~\cite{nie2025large, austin2021structured} define a forward noising process and learn a reverse denoising process for generation.
Multimodal dLLMs~\cite{you2025llada, li2025lavida, yang2025mmada} extend this framework by conditioning on visual input: a frozen vision encoder maps the input image to $N_v$ visual tokens $\mathbf{v}$, which, together with $N_c$ instruction tokens $\mathbf{c}$, form a fixed prefix exempt from masking.
The forward process acts exclusively on the $N_g$ generation tokens $\mathbf{x}_0$, independently masking each with probability $t \in [0,1]$:
\begin{equation}
    q_{t|0}(\mathbf{x}_t \mid \mathbf{x}_0)
    = \prod_{i=1}^{N_g}
      \mathrm{Cat}\!\left(x^i_t;\;
      (1\!-\!t)\,\delta_{x^i_0} + t\,\delta_{[\textsc{mask}]}\right).
    \label{eq:forward}
\end{equation}
The model $p_\theta(x^i_0 \mid \mathbf{x}_t, \mathbf{v}, \mathbf{c})$ is trained to predict the clean token at each masked position.
At inference, generation proceeds from a fully masked sequence through $T$ discrete steps; at each step, the model predicts all masked positions and selects ${\sim}N_g/T$ to unmask based on prediction confidence, with previously revealed tokens preserved.
Each step processes the concatenated sequence $[\mathbf{v};\, \mathbf{c};\, \mathbf{x}_t]$ of length $N = N_v + N_c + N_g$ through $L$ layers of bidirectional self-attention with $H$ heads, yielding attention tensors $\{\mathbf{A}^{(t)}_l \in \mathbb{R}^{H \times N \times N}\}_{l=1}^{L}$.
The total inference cost thus scales as $\mathcal{O}(T \cdot L \cdot N^2)$, where $N_v$ dominates the sequence length.
Since $\mathbf{v}$ is fixed across all $T$ steps, its quadratic attention cost is replicated at every step without diminishing return.
\subsection{Multimodal Diffusion Language Models.}
Multimodal dLLMs~\cite{you2025llada, li2025lavida, yang2025mmada} extend this framework by conditioning generation on a visual input.
A frozen vision encoder maps an input image to $N_v$ visual tokens $\mathbf{v} = (v^1, \ldots, v^{N_v})$; together with $N_c$ instruction tokens $\mathbf{c} = (c^1, \ldots, c^{N_c})$, these form a fixed conditioning prefix that is never subject to masking.
The forward and reverse processes act exclusively on the $N_g$ generation tokens $\mathbf{x}_t$, with the model's predictive distribution conditioned on the full context: $p_\theta(x^i_0 \mid \mathbf{x}_t, \mathbf{v}, \mathbf{c})$.
At each denoising step, the model processes the concatenated sequence $[\mathbf{v};\, \mathbf{c};\, \mathbf{x}_t]$ of length $N = N_v + N_c + N_g$ through $L$ layers of bidirectional self-attention with $H$ heads, yielding attention tensors $\bigl\{\mathbf{A}^{(t)}_l \in \mathbb{R}^{H \times N \times N}\bigr\}_{l=1}^{L}$.
The total inference cost scales as $\mathcal{O}(T \cdot L \cdot N^2)$, where $N_v$ constitutes the dominant factor in sequence length.
As $\mathbf{v}$ is fixed across all $T$ steps, its quadratic attention cost is replicated at every step without diminishing return, leading to a structural inefficiency that does not arise in text-only dLLMs.



\end{document}